\documentclass[10pt,twocolumn,letterpaper]{article}

\usepackage{cvpr}
\usepackage{times}
\usepackage{epsfig}
\usepackage{graphicx}
\usepackage{amsmath}
\usepackage{amssymb}
\usepackage[export]{adjustbox}
\usepackage{makecell}
\usepackage{siunitx}
\cvprfinalcopy 

\iftrue  

\newcommand{\TODO}[1]{}
\newcommand{\outline}[1]{}
\newcommand{\textgray}[1]{}
\newcommand{\commenttext}[1]{}
\newcommand{\commentfoot}[1]{}
\newcommand{\commentselfoot}[2]{}
\newcommand{\commentselrep}[2]{}
\newcommand{\topic}[1]{}
\else 
\newcommand{\TODO}[1]{{\textcolor{red}{[[TODO: {#1}]]}}}
\newcommand{\outline}[1]{{\textcolor{blue}{[[{#1}]]}}}
\newcommand{\textgray}[1]{\textcolor{gray}{[[{#1}]]}}
\newcommand{\commenttext}[1]{\textcolor{red}{[[{#1}]]}}
\newcommand{\commentfoot}[1]{\footnote{\textcolor{red}{\textit{#1}}}}
\newcommand{\commentselfoot}[2]{{\textcolor{blue}{#1}}\commenttext{#2}}
\newcommand{\commentselrep}[2] {{\textcolor{blue}{#1}} {\textcolor{green}{[[\textit{#2}]]}}}
\newcommand{\topic}[1]{\textcolor{gray}{\textbf{(#1.)}}}
\fi

\newcommand{\ignore}[1]{}

\def\cvprPaperID{1539} 

\ifcvprfinal\fi
\begin{document}

\title{Object Contour Detection with a Fully Convolutional Encoder-Decoder Network}
\author{Jimei Yang\\
Adobe Research\\
{\tt\small jimyang@adobe.com}
\and
Brian Price\\
Adobe Research\\
{\tt\small bprice@adobe.com}
\and
Scott Cohen\\
Adobe Research\\
{\tt\small scohen@adobe.com}
\and
Honglak Lee\\
University of Michigan, Ann Arbor\\
{\tt\small honglak@umich.edu}
\and
Ming-Hsuan Yang\\
UC Merced\\
{\tt\small mhyang@ucmerced.edu}
}

\maketitle

\begin{abstract}
We develop a deep learning algorithm for contour detection with a fully convolutional encoder-decoder network.
Different from previous low-level edge detection, our algorithm focuses on detecting higher-level object contours.
%
%
Our network is trained end-to-end on PASCAL VOC with refined ground truth from inaccurate polygon annotations, yielding much higher precision in object contour detection than previous methods.
We find that the learned model generalizes well to unseen object classes from the same super-categories on MS COCO and can match state-of-the-art edge detection on BSDS500 with fine-tuning.
By combining with the multiscale combinatorial grouping algorithm, our method can generate high-quality segmented object proposals, 
which significantly advance the state-of-the-art on PASCAL VOC (improving average recall from 0.62 to 0.67) with a relatively small amount of candidates ($\sim$1660 per image).
\vspace{-2mm}
\end{abstract}

\section{Introduction}
\begin{figure}[t]
\centering
\setlength{\tabcolsep}{1pt}
\small{
\begin{tabular}{ccc}
\includegraphics[width=0.3\linewidth,keepaspectratio,cframe=black 1pt]{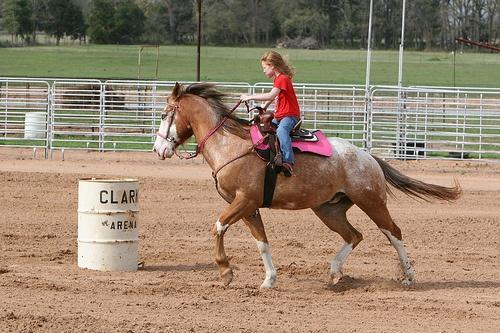} &
\includegraphics[width=0.3\linewidth,keepaspectratio,cframe=black 1pt]{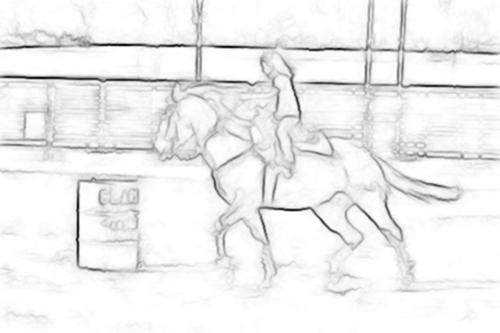} &
\includegraphics[width=0.3\linewidth,keepaspectratio,cframe=black 1pt]{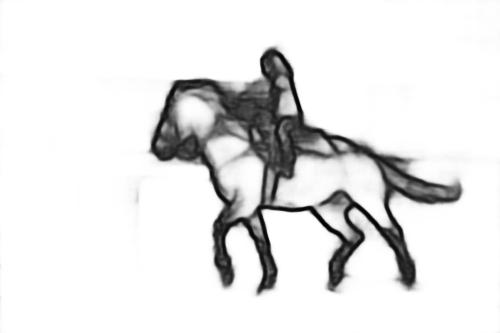} \\
\includegraphics[width=0.3\linewidth,keepaspectratio,cframe=black 1pt]{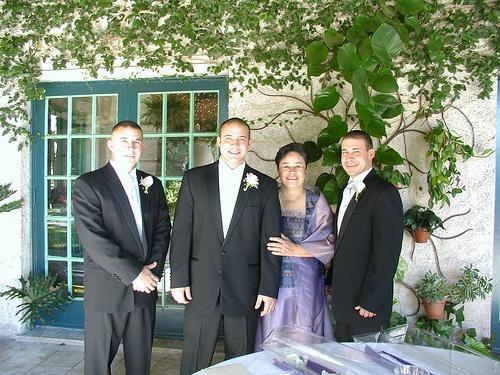} &
\includegraphics[width=0.3\linewidth,keepaspectratio,cframe=black 1pt]{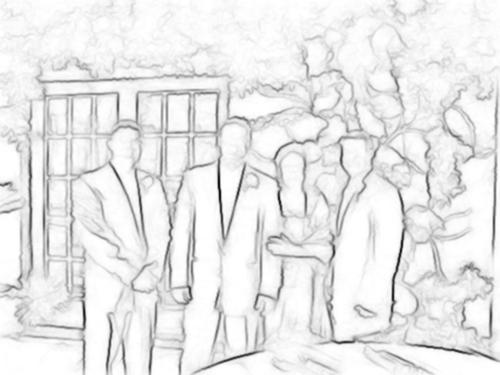} &
\includegraphics[width=0.3\linewidth,keepaspectratio,cframe=black 1pt]{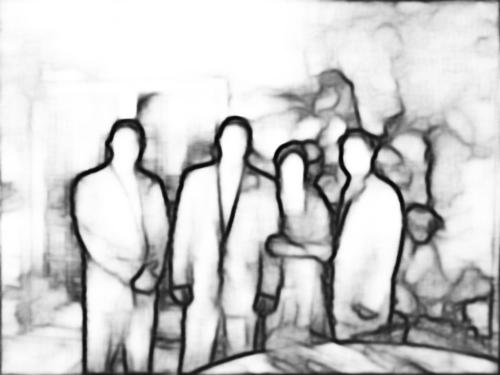} \\
(a) Image & (b) \cite{Dollar_ICCV_2013} & Ours \\
\end{tabular}
}
\caption{Object contour detection. Given input images (a), our model can effectively learn to detect contours of foreground objects (c) in contrast to traditional edge detection (b).}
\label{fig:intro}
\vspace{-3mm}
\end{figure}

Object contour detection is fundamental for numerous vision tasks.
For example, it can be used for image segmentation~\cite{Rother_SIGGRAPH_2004, Arbelaez_PAMI_2011}, for object detection~\cite{Ferrari_PAMI_2008,Girshick_CVPR_2014}, and for occlusion and depth reasoning~\cite{Hoiem_ICCV_2007, Amer_IJCV_2015}.
Given its axiomatic importance, however, we find that object contour detection is relatively under-explored in the literature.
At the same time, many works have been devoted to edge detection that responds to both foreground objects and background boundaries (Figure~\ref{fig:intro} (b)).
In this paper, we address ``object-only" contour detection that is expected to suppress background boundaries (Figure~\ref{fig:intro}(c)).

Edge detection has a long history. 
Early research focused on designing simple filters to detect pixels with highest gradients in their local neighborhood, e.g. Sobel~\cite{Forsyth_CV_2003} and Canny~\cite{Canny_PAMI_1986}.
The main problem with filter based methods is that they only look at the color or brightness differences between adjacent pixels but cannot tell the texture differences in a larger receptive field.
With the advance of texture descriptors~\cite{Malik_IJCV_2001}, Martin et al.~\cite{Martin_PAMI_2004} combined color, brightness and texture gradients in their probabilistic boundary detector.
Arbelaez et al.~\cite{Arbelaez_PAMI_2011} further improved upon this by computing local cues from multiscale and spectral clustering, known as \emph{gPb}, which yields state-of-the-art accuracy.
However, the globalization step of \emph{gPb} significantly increases the computational load.
Lim and Dollar~\cite{Lim_CVPR_2013, Dollar_ICCV_2013} analyzed the clustering structure of local contour maps and developed efficient supervised learning algorithms for fast edge detection~\cite{Dollar_ICCV_2013}.
These efforts lift edge detection to a higher abstract level, but still fall below human perception due to their lack of object-level knowledge.

Recently deep convolutional networks~\cite{Krizhevsky_NIPS_2012} have demonstrated remarkable ability of learning high-level representations for object recognition~\cite{Girshick_CVPR_2014, Chen_arXiv_2014}. 
These learned features have been adopted to detect natural image edges~\cite{Kivinen_AISTATS_2014, Bertasius_CVPR_2015,Shen_CVPR_2015, Xie_ICCV_2015} and yield a new state-of-the-art performance~\cite{Xie_ICCV_2015}.
All these methods require training on ground truth contour annotations.
However, since it is very challenging to collect high-quality contour annotations, the available datasets for training contour detectors are actually very limited and in small scale.
For example, the standard benchmarks, Berkeley segmentation (BSDS500)~\cite{Martin_ICCV_2001} and NYU depth v2 (NYUDv2)~\cite{Silberman_ECCV_2012} datasets only include 200 and 381 training images, respectively.
Therefore, the representation power of deep convolutional networks has not been entirely harnessed for contour detection.
In this paper, we scale up the training set of deep learning based contour detection to more than 10k images on PASCAL VOC~\cite{PASCAL_VOC_2010}. 
To address the quality issue of ground truth contour annotations, we develop a dense CRF~\cite{Krahenbuhl_NIPS_2011} based method to refine the object segmentation masks from polygons.

Given image-contour pairs, we formulate object contour detection as an image labeling problem. 
Inspired by the success of fully convolutional networks~\cite{Long_CVPR_2015} and deconvolutional networks~\cite{Noh_ICCV_2015} on semantic segmentation, 
we develop a fully convolutional encoder-decoder network (CEDN).
Being fully convolutional, our CEDN network can operate on arbitrary image size and the encoder-decoder network emphasizes its asymmetric structure that differs from deconvolutional network~\cite{Noh_ICCV_2015}.
We initialize our encoder with VGG-16 net~\cite{Simonyan_ICLR_2015} (up to the ``fc6" layer) and to achieve dense prediction of image size our decoder is constructed by alternating unpooling and convolution layers where unpooling layers re-use the switches from max-pooling layers of encoder to upscale the feature maps.
During training, we fix the encoder parameters (VGG-16) and only optimize decoder parameters. 
This allows the encoder to maintain its generalization ability so that the learned decoder network can be easily combined with other tasks, such as bounding box regression or semantic segmentation. 

We evaluate the trained network on unseen object categories from BSDS500 and MS COCO datasets~\cite{Lin_COCO_2014}, 
and find the network generalizes well to objects in similar ``super-categories'' to those in the training set, e.g. it generalizes to objects like ``bear" in the ``animal'' super-category since ``dog'' and ``cat" are in the training set.
We also show the trained network can be easily adapted to detect natural image edges through a few iterations of fine-tuning, which produces comparable results with the state-of-the-art algorithm~\cite{Xie_ICCV_2015}.

An immediate application of contour detection is generating object proposals.
Previous literature has investigated various methods of generating bounding box or segmented object proposals by scoring edge features~\cite{Zitnick_ECCV_2014, Cheng_CVPR_2014} and combinatorial grouping~\cite{deSande_ICCV_2011,Carreira_CVPR_2010, Arbelaez_CVPR_2014} and etc.
In this paper, we use a multiscale combinatorial grouping (MCG) algorithm~\cite{Arbelaez_CVPR_2014} to generate segmented object proposals from our contour detection.
As a result, our method significantly improves the quality of segmented object proposals on the PASCAL VOC 2012 validation set, achieving 0.67 average recall from overlap 0.5 to 1.0 with only about 1660 candidates per image, compared to the state-of-the-art average recall 0.62 by original \emph{gPb}-based MCG algorithm with near 5140 candidates per image.
We also evaluate object proposals on the MS COCO dataset with 80 object classes and analyze the average recalls from different object classes and their super-categories.

The key contributions are summarized below:
\begin{itemize}
\item We develop a simple yet effective fully convolutional encoder-decoder network for object contour prediction and the trained model generalizes well to unseen object classes from the same super-categories, yielding significantly higher precision in object contour detection than previous methods.
\item We show we can fine tune our network for edge detection and match the state-of-the-art in terms of precision and recall.
\item We generate accurate object contours from imperfect polygon based segmentation annotations, which makes it possible to train an object contour detector at scale.
\item Our method obtains state-of-the-art results on segmented object proposals by integrating with combinatorial grouping~\cite{Arbelaez_CVPR_2014}.
\end{itemize}

\section{Related Work}
\paragraph{Semantic contour detection.}
Hariharan et al.~\cite{Hariharan_ICCV_2011} study top-down contour detection problem.
Their semantic contour detectors~\cite{Hariharan_ICCV_2011} are devoted to find the semantic boundaries between different object classes.
Although they consider object instance contours while collecting annotations, they choose to ignore the occlusion boundaries between object instances from the same class. 
%
%
As combining bottom-up edges with object detector output, their method can be extended to object instance contours but might encounter challenges of generalizing to unseen object classes. 
\paragraph{Occlusion boundary detection.}
%
%
Hoiem et al.~\cite{Hoiem_ICCV_2007} study the problem of recovering occlusion boundaries from a single image.
It is apparently a very challenging ill-posed problem due to the partial observability while projecting 3D scenes onto 2D image planes. 
They formulate a CRF model to integrate various cues: color, position, edges, surface orientation and depth estimates.
We believe our instance-level object contours will provide another strong cue for addressing this problem that is worth investigating in the future.
\paragraph{Object proposal generation.}
There is a large body of works on generating bounding box or segmented object proposals.
Hosang et al.~\cite{Hosang_PAMI_2015} and Jordi et al.~\cite{Pont_Tuset_ICCV_2015} present nice overviews and analyses about the state-of-the-art algorithms.
Bounding box proposal generation~\cite{deSande_ICCV_2011, Zitnick_ECCV_2014, Cheng_CVPR_2014, Alexe_PAMI_2012} is motivated by efficient object detection. 
One of their drawbacks is that bounding boxes usually cannot provide accurate object localization.
More related to our work is generating segmented object proposals~\cite{Arbelaez_CVPR_2014, Carreira_CVPR_2010, Endres_ECCV_2010, Humayun_CVPR_2014, Kim_ECCV_2012, Krahenbuhl_ECCV_2014, Rantalankila_CVPR_2014}.
At the core of segmented object proposal algorithms is contour detection and superpixel segmentation. 
We experiment with a state-of-the-art method of multiscale combinatorial grouping~\cite{Arbelaez_CVPR_2014} to generate proposals and believe our object contour detector can be directly plugged into most of these algorithms.

\section{Object Contour Detection}
In this section, we introduce our object contour detection method with the proposed fully convolutional encoder-decoder network.
\subsection{Fully Convolutional Encoder-Decoder Network}
We formulate contour detection as a binary image labeling problem where ``1" and ``0" indicates ``contour" and ``non-contour", respectively.
Image labeling is a task that requires both high-level knowledge and low-level cues.
Given the success of deep convolutional networks~\cite{Krizhevsky_NIPS_2012} for learning rich feature hierarchies,
image labeling has been greatly advanced, especially on the task of semantic segmentation~\cite{Chen_arXiv_2014,Long_CVPR_2015,Sifei_CVPR_2015,CRFasRNN_ICCV_2015,Noh_ICCV_2015,Liu_ICCV_2015}. 
Among those end-to-end methods, fully convolutional networks~\cite{Long_CVPR_2015} scale well up to the image size but cannot produce very accurate labeling boundaries; unpooling layers help deconvolutional networks~\cite{Noh_ICCV_2015} to generate better label localization but their symmetric structure introduces a heavy decoder network which is difficult to train with limited samples.
\begin{figure*}[tbhp]
\centering
\includegraphics[width=0.9\linewidth,keepaspectratio]{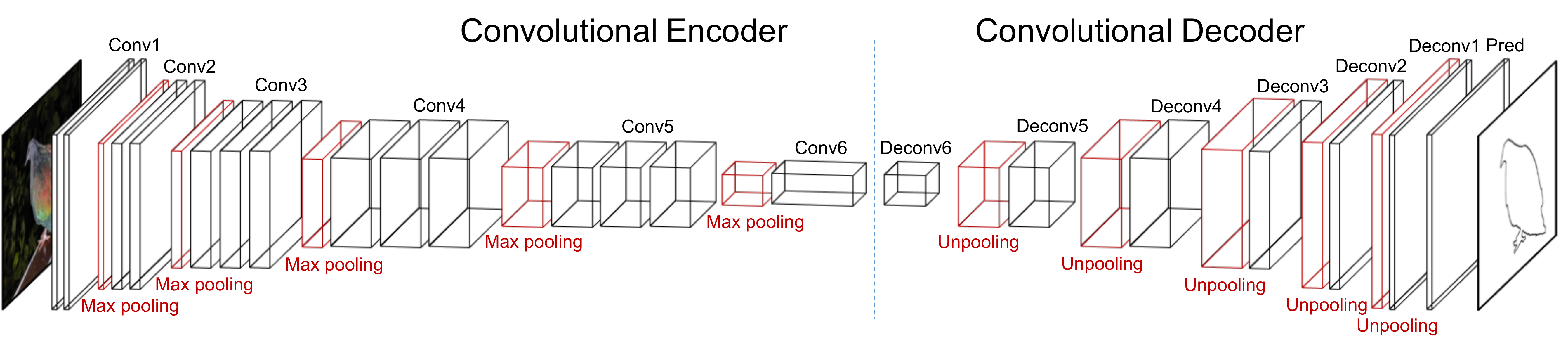}
\caption{Architecture of the proposed fully convolutional encoder-decoder network.}
\label{fig:cedn}
\end{figure*}

We borrow the ideas of full convolution and unpooling from above two works and develop a fully convolutional encoder-decoder network for object contour detection. 
The network architecture is demonstrated in Figure~\ref{fig:cedn}.
We use the layers up to ``fc6" from VGG-16 net~\cite{Simonyan_ICLR_2015} as our encoder.
Since we convert the ``fc6" to be convolutional, so we name it ``conv6" in our decoder.

Due to the asymmetric nature of image labeling problems (image input and mask output), we break the symmetric structure of deconvolutional networks and introduce a light-weighted decoder. 
The first layer of decoder ``deconv6" is designed for dimension reduction that projects 4096-d ``conv6" to 512-d with $1\!\times\!1$ kernel so that we can re-use the pooling switches from ``conv5" to upscale the feature maps by twice in the following ``deconv5" layer. 
The number of channels of every decoder layer is properly designed to allow unpooling from its corresponding max-pooling layer.
All the decoder convolution layers except ``deconv6" use $5\!\times\!5$ kernels.
All the decoder convolution layers except the one next to the output label are followed by relu activation function.
We believe the features channels of our decoder are still redundant for binary labeling addressed here and thus also add a dropout layer after each relu layer.
A complete decoder network setup is listed in Table~\ref{tab:decoder} and the loss function is simply the pixel-wise logistic loss.
\begin{table}[!htb]
\centering
\caption{Decoder network setup.}
\label{tab:decoder}
\setlength{\tabcolsep}{1pt}
\begin{tabular}{c|c|c|c|c}
\hline
name & deconv6 & deconv5 & deconv4 &\\
\Xhline{4\arrayrulewidth}
setup & conv & unpool-conv & unpool-conv & \\
kernel & $1\!\times\!1\!\times\!512$ & $5\!\times\!5\!\times\!512$ & $5\!\times\!5\!\times\!256$ \\
acti & relu & relu & relu & \\
\hline
name & deconv3 & deconv2 & deconv1 & pred\\
\Xhline{4\arrayrulewidth}
setup & unpool-conv & unpool-conv & unpool-conv & conv \\
kernel & $5\!\times\!5\!\times\!128$ & $5\!\times\!5\!\times\!64$ & $5\!\times\!5\!\times\!32$ & $5\!\times\!5\!\times\!1$\\
activation & relu & relu & relu & sigmoid \\
\hline
\end{tabular}
\vspace{-2mm}
\end{table}

\subsection{Contour Ground Truth Refinement}
Drawing detailed and accurate contours of objects is a challenging task for human beings. 
This is why many large scale segmentation datasets~\cite{Russell_IJCV_2008,PASCAL_VOC_2010,Lin_COCO_2014} provide contour annotations with polygons as they are less expensive to collect at scale.
However, because of unpredictable behaviors of human annotators and limitations of polygon representation, the annotated contours usually do not align well with the true image boundaries and thus cannot be directly used as ground truth for training. 
Among all, the PASCAL VOC dataset is a widely-accepted benchmark with high-quality annotation for object segmentation. 
VOC 2012 release includes 11540 images from 20 classes covering a majority of common objects from categories such as ``person", ``vehicle", ``animal" and ``household", where 1464 and 1449 images are annotated with object instance contours for training and validation.
Hariharan et al.~\cite{Hariharan_ICCV_2011} further contribute more than 10000 high-quality annotations to the remaining images. 
Together there are 10582 images for training and 1449 images for validation (the exact 2012 validation set).
We choose this dataset for training our object contour detector with the proposed fully convolutional encoder-decoder network.

\begin{figure}[tbhp]
\centering
\setlength{\tabcolsep}{1pt}
\small{
\begin{tabular}{cc}
\includegraphics[width=0.45\linewidth,keepaspectratio]{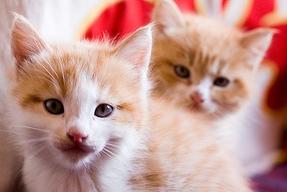} &
\includegraphics[width=0.45\linewidth,keepaspectratio]{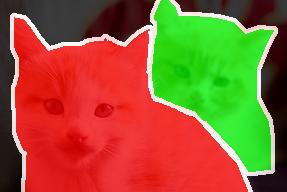} \\
(a) Image & (b) Annotation \\
\includegraphics[width=0.45\linewidth,keepaspectratio]{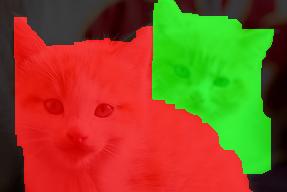} &
\includegraphics[width=0.45\linewidth,keepaspectratio]{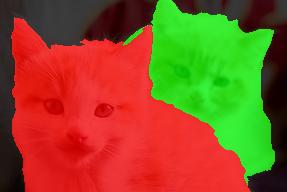} \\
(c) GraphCut refinement & (d) DenseCRF refinement
\end{tabular}
}
\caption{Contour refinement. The polygon based annotations (a) cannot be directly used for training due to its inaccurate boundaries (thin white area reflects unlabeled pixels between objects). We align them to image boundaries by re-labeling the uncertain areas with dense CRF (d), compared to Graph Cut (c).}
\label{fig:contour_refinement}
\vspace{-2mm}
\end{figure}

The original PASCAL VOC annotations leave a thin unlabeled (or uncertain) area between occluded objects (Figure~\ref{fig:contour_refinement}(b)).
To find the high-fidelity contour ground truth for training, we need to align the annotated contours with the true image boundaries.
We consider contour alignment as a multi-class labeling problem and introduce a dense CRF model~\cite{Krahenbuhl_NIPS_2011} where every instance (or background) is assigned with one unique label. 
%
%
The dense CRF optimization then fills the uncertain area with neighboring instance labels so that we obtain refined contours at the labeling boundaries (Figure~\ref{fig:contour_refinement}(d)).
We also experimented with the Graph Cut method~\cite{Boykov_ICCV_2001} but find it usually produces jaggy contours due to its shortcutting bias (Figure~\ref{fig:contour_refinement}(c)).

\subsection{Training}
We train the network using \emph{Caffe}~\cite{Jia_Caffe_2014}.
For each training image, we randomly crop four $224\!\times\!224\!\times\!3$ patches and together with their mirrored ones compose a $224\!\times\!224\!\times\!3\!\times\!8$ minibatch.
The ground truth contour mask is processed in the same way.
We initialize the encoder with pre-trained VGG-16 net and the decoder with random values. 
During training, we fix the encoder parameters and only optimize the decoder parameters. 
This allows our model to be easily integrated with other decoders such as bounding box regression~\cite{Girshick_ICCV_2015} and semantic segmentation~\cite{Noh_ICCV_2015} for joint training. 
As the ``contour" and ``non-contour" pixels are extremely imbalanced in each minibatch, the penalty for being ``contour" is set to be 10 times the penalty for being ``non-contour".
We use the Adam method~\cite{Ba_ICLR_2015} to optimize the network parameters and find it is more efficient than standard stochastic gradient descent.
We set the learning rate to $10^{-4}$ and train the network with 30 epochs with all the training images being processed each epoch.
Note that we fix the training patch to $224\!\times\!224$ for memory efficiency and the learned parameters can be used on images of arbitrary size because of its fully convolutional nature.

\section{Results}
In this section, we evaluate our method on contour detection and proposal generation using three datasets: PASCAL VOC 2012, BSDS500 and MS COCO.
We will explain the details of generating object proposals using our method after the contour detection evaluation.
More evaluation results are in the supplementary materials.
\subsection{Contour Detection}
Given trained models, all the test images are fed-forward through our CEDN network in their original sizes to produce contour detection maps.
The detection accuracies are evaluated by four measures: F-measure (F), fixed contour threshold (ODS), per-image best threshold (OIS) and average precision (AP).
Note that a standard non-maximum suppression is used to clean up the predicted contour maps (thinning the contours) before evaluation.
\begin{figure}[h]
\centering
\includegraphics[width=1.0\linewidth,keepaspectratio]{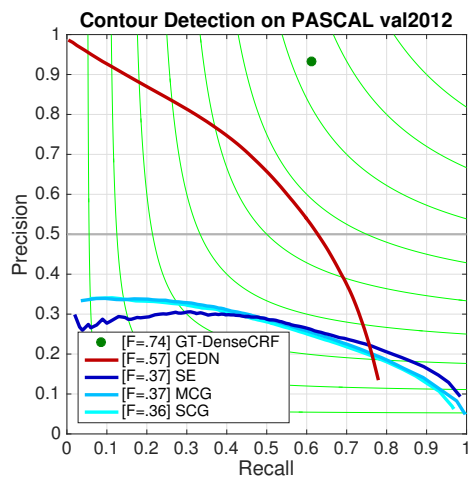}
\caption{PR curve for contour detection on the PASCAL VOC 2012 validation set.}
\label{fig:pr_curve_pascal2012}
\vspace{-4mm}
\end{figure}
\paragraph{PASCAL val2012.} 
\begin{figure*}[tbhp]
\centering
\setlength{\tabcolsep}{1pt}
\begin{tabular}{ccccc}
\includegraphics[width=0.18\linewidth,keepaspectratio,cframe=black 1pt]{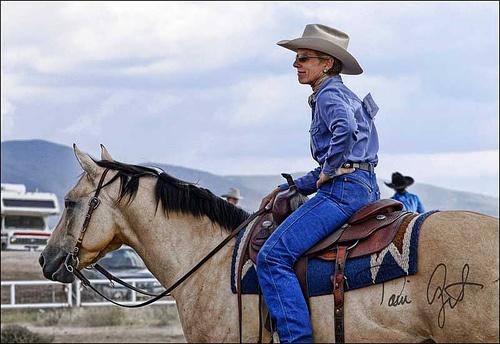} &
\includegraphics[width=0.18\linewidth,keepaspectratio,cframe=black 1pt]{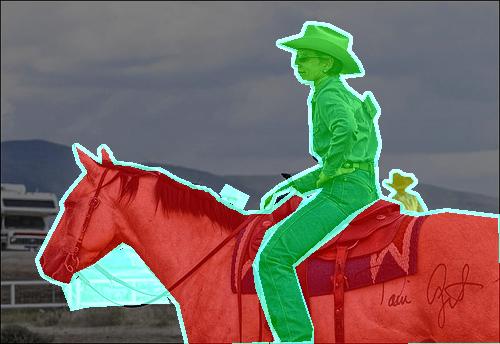} &
\includegraphics[width=0.18\linewidth,keepaspectratio,cframe=black 1pt]{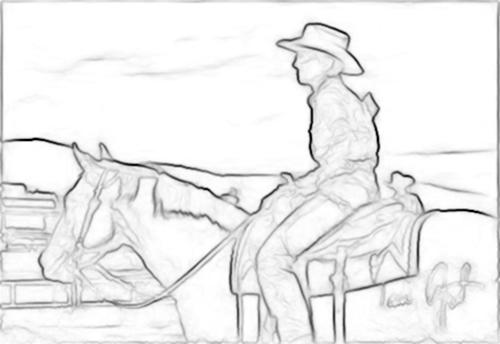} &
\includegraphics[width=0.18\linewidth,keepaspectratio,cframe=black 1pt]{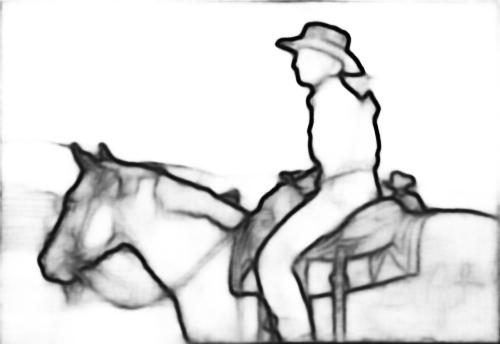} &
\includegraphics[width=0.18\linewidth,keepaspectratio,cframe=black 1pt]{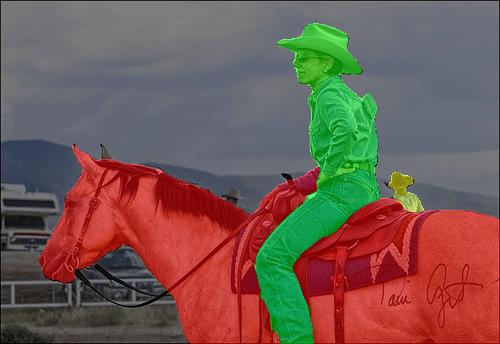} \\
\includegraphics[width=0.18\linewidth,keepaspectratio,cframe=black 1pt]{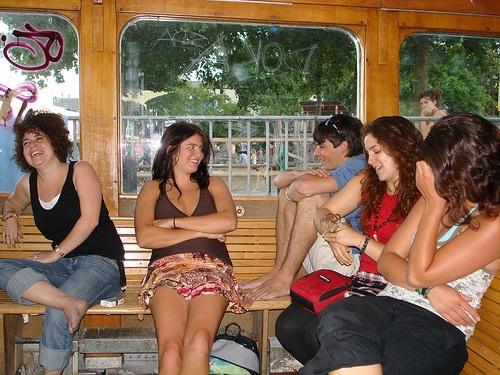} &
\includegraphics[width=0.18\linewidth,keepaspectratio,cframe=black 1pt]{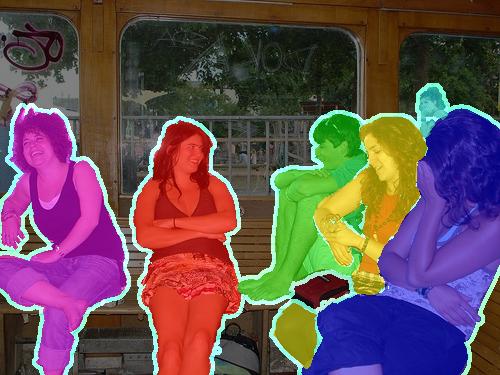} &
\includegraphics[width=0.18\linewidth,keepaspectratio,cframe=black 1pt]{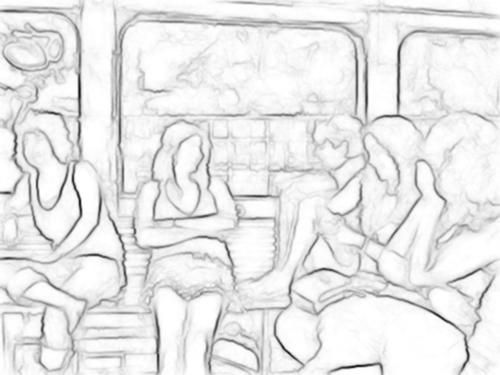} &
\includegraphics[width=0.18\linewidth,keepaspectratio,cframe=black 1pt]{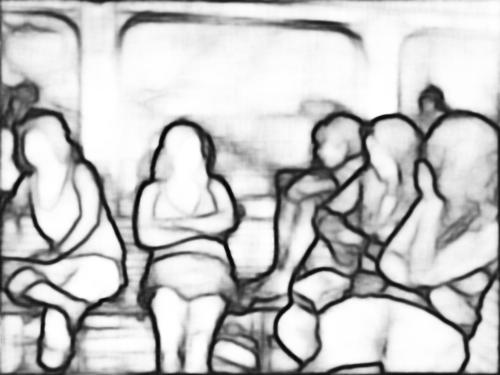} &
\includegraphics[width=0.18\linewidth,keepaspectratio,cframe=black 1pt]{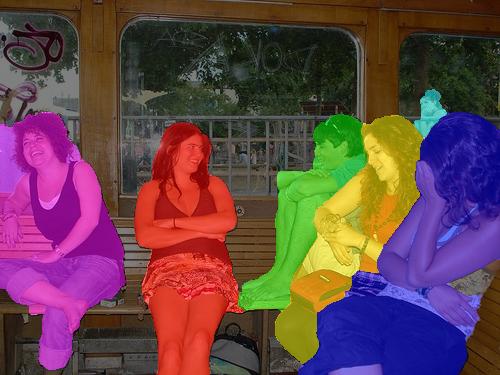} \\
\includegraphics[width=0.18\linewidth,keepaspectratio,cframe=black 1pt]{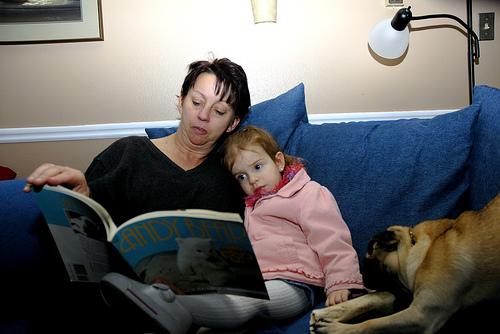} &
\includegraphics[width=0.18\linewidth,keepaspectratio,cframe=black 1pt]{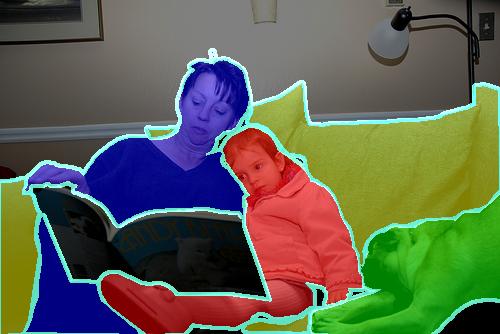} &
\includegraphics[width=0.18\linewidth,keepaspectratio,cframe=black 1pt]{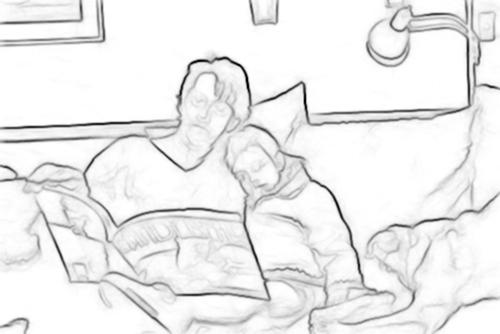} &
\includegraphics[width=0.18\linewidth,keepaspectratio,cframe=black 1pt]{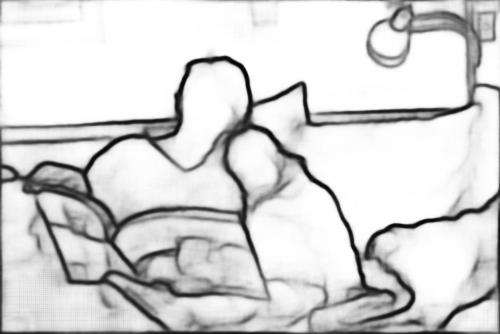} &
\includegraphics[width=0.18\linewidth,keepaspectratio,cframe=black 1pt]{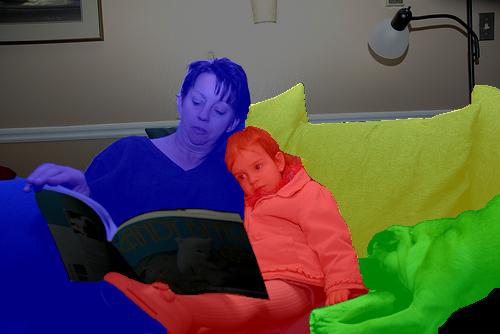} \\
\includegraphics[width=0.18\linewidth,keepaspectratio,cframe=black 1pt]{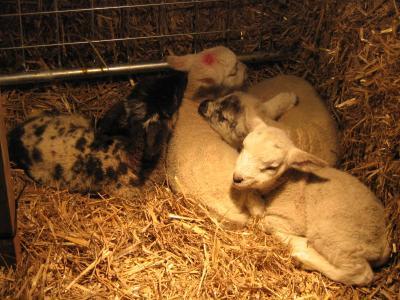} &
\includegraphics[width=0.18\linewidth,keepaspectratio,cframe=black 1pt]{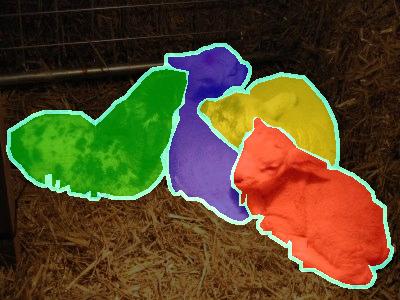} &
\includegraphics[width=0.18\linewidth,keepaspectratio,cframe=black 1pt]{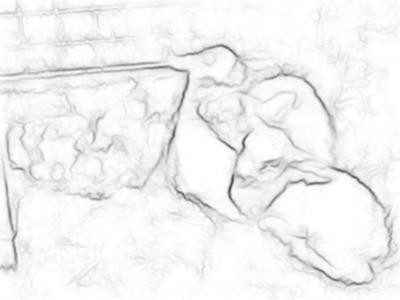} &
\includegraphics[width=0.18\linewidth,keepaspectratio,cframe=black 1pt]{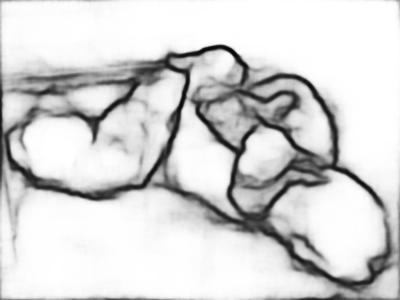} &
\includegraphics[width=0.18\linewidth,keepaspectratio,cframe=black 1pt]{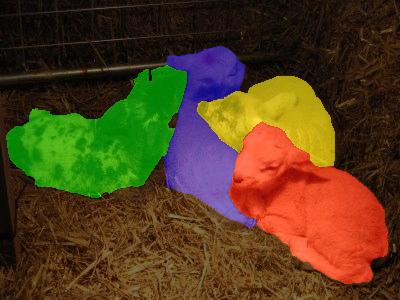} \\
(a) & (b)  & (c) & (d) & (e) \\
\end{tabular}
\caption{Example results on PASCAL VOC val2012. In each row from left to right we present (a) input image, (b) ground truth annotation, (c) edge detection~\cite{Dollar_ICCV_2013}, (d) our object contour detection and (e) our best object proposals.}
\label{fig:pascal_example}
\vspace{-2mm}
\end{figure*}
\ignore{
\begin{figure*}[tbhp]
\centering
\setlength{\tabcolsep}{1pt}
\begin{tabular}{cccc}
\includegraphics[width=0.23\linewidth,keepaspectratio,cframe=black 1pt]{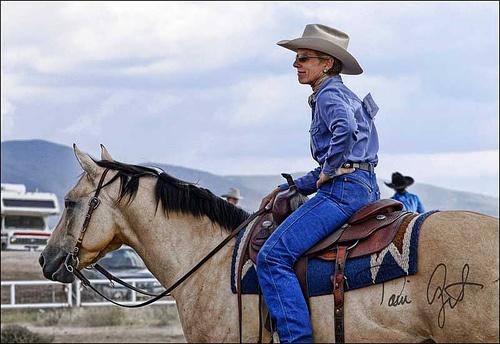} &
\includegraphics[width=0.23\linewidth,keepaspectratio,cframe=black 1pt]{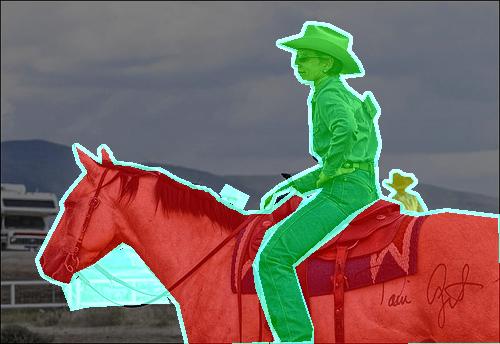} &
\includegraphics[width=0.23\linewidth,keepaspectratio,cframe=black 1pt]{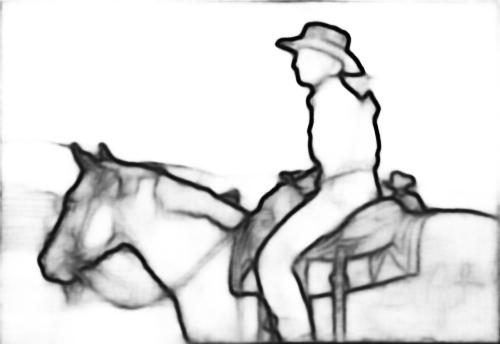} &
\includegraphics[width=0.23\linewidth,keepaspectratio,cframe=black 1pt]{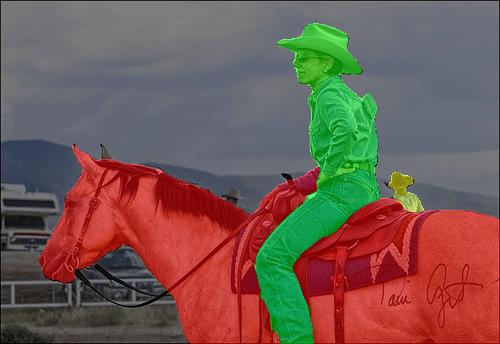} \\
\includegraphics[width=0.23\linewidth,keepaspectratio,cframe=black 1pt]{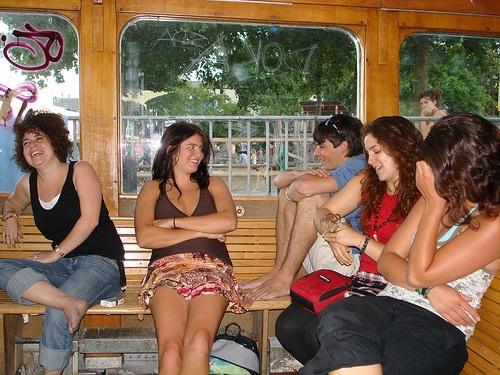} &
\includegraphics[width=0.23\linewidth,keepaspectratio,cframe=black 1pt]{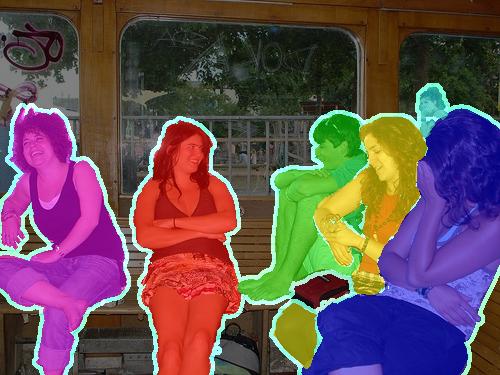} &
\includegraphics[width=0.23\linewidth,keepaspectratio,cframe=black 1pt]{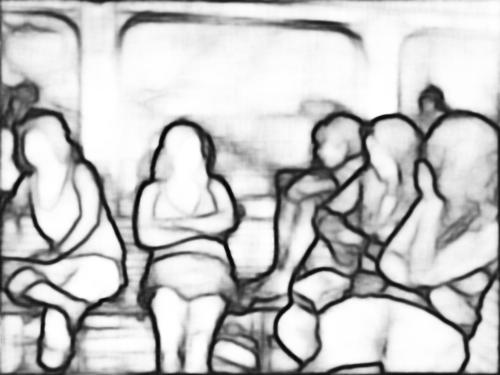} &
\includegraphics[width=0.23\linewidth,keepaspectratio,cframe=black 1pt]{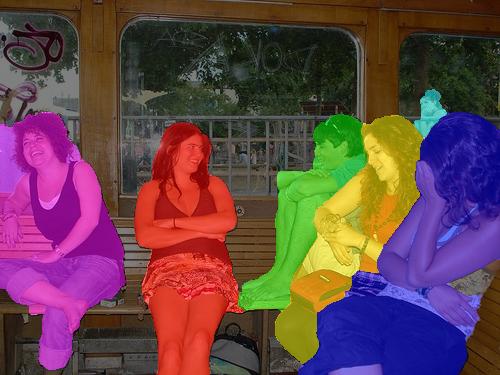} \\
\includegraphics[width=0.23\linewidth,keepaspectratio,cframe=black 1pt]{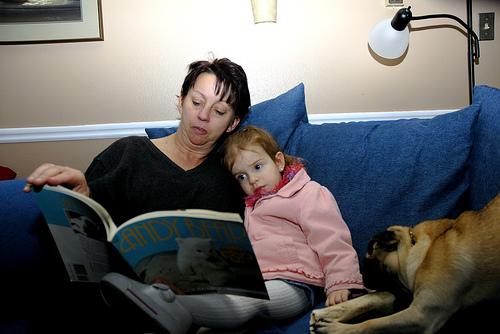} &
\includegraphics[width=0.23\linewidth,keepaspectratio,cframe=black 1pt]{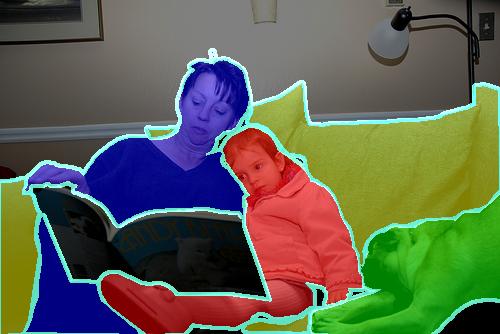} &
\includegraphics[width=0.23\linewidth,keepaspectratio,cframe=black 1pt]{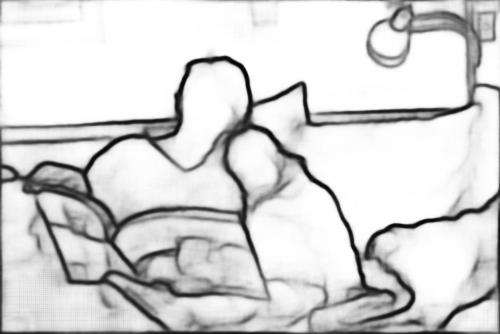} &
\includegraphics[width=0.23\linewidth,keepaspectratio,cframe=black 1pt]{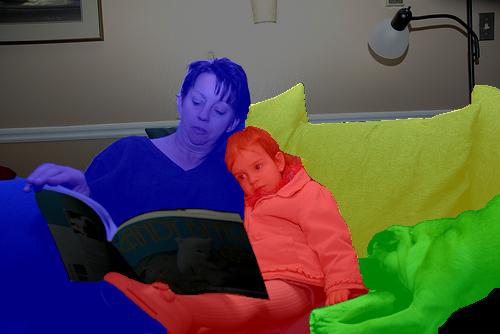} \\
(a) & (b)  & (c) & (d) \\
\end{tabular}
\caption{Example results on PASCAL VOC val2012. In each row from left to right we present (a) input image, (b) ground truth annotation, (c) contour detection and (d) best object proposals.}
\label{fig:pascal_example}
\vspace{-2mm}
\end{figure*}
}
We first present results on the PASCAL VOC 2012 validation set, shortly ``PASCAL val2012", with comparisons to three baselines, structured edge detection (SE)~\cite{Dollar_ICCV_2013}, singlescale combinatorial grouping (SCG) and multiscale combinatorial grouping (MCG)~\cite{Arbelaez_CVPR_2014}.
Precision-recall curves are shown in Figure~\ref{fig:pr_curve_pascal2012}.
Note that we use the originally annotated contours instead of our refined ones as ground truth for unbiased evaluation.
Accordingly we consider the refined contours as the upper bound since our network is learned from them.
Its precision-recall value is referred as ``GT-DenseCRF" with a green spot in Figure~\ref{fig:pr_curve_pascal2012}.
Compared to the baselines, our method (CEDN) yields very high precisions, which means it generates visually cleaner contour maps with background clutters well suppressed (the third column in Figure~\ref{fig:pascal_example}).
Note that the occlusion boundaries between two instances from the same class are also well recovered by our method (the second example in Figure~\ref{fig:pascal_example}).
We also note that there is still a big performance gap between our current method (F=0.57) and the upper bound (F=0.74), which requires further research for improvement.
\ignore{
\begin{table}[!htb]
\centering
\caption{Contour detection results on PASCAL val2012.}
\label{tab:pascal2012}
\begin{tabular}{c|c|c|c}
\hline
 & ODS & OIS & AP \\
\Xhline{4\arrayrulewidth}
SCG &  .360 & .427 & .235 \\
MCG &  .366 & .437 & .248 \\
\hline
SE & .357 & .399 & .247 \\
\hline
CEDN & .606 & .630 & .590 \\
\hline
\end{tabular}
\vspace{-2mm}
\end{table}
}
\begin{figure}[tbhp]
\vspace{-2mm}
\centering
\setlength{\tabcolsep}{1pt}
\begin{tabular}{cccc}
\includegraphics[width=0.23\linewidth,keepaspectratio,cframe=black 1pt]{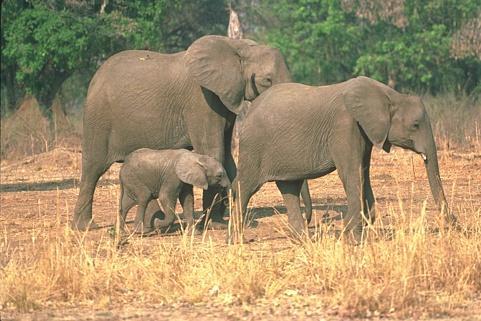} &
\includegraphics[width=0.23\linewidth,keepaspectratio,cframe=black 1pt]{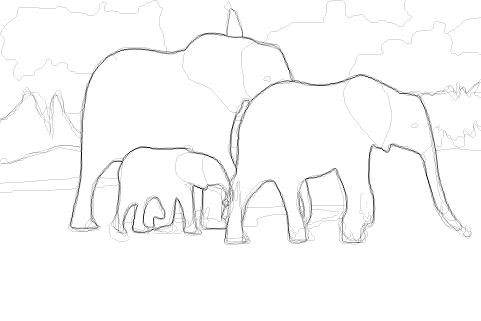} &
\includegraphics[width=0.23\linewidth,keepaspectratio,cframe=black 1pt]{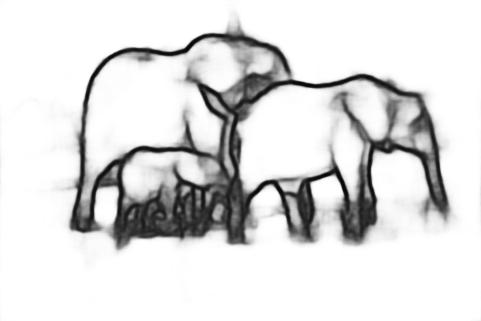} &
\includegraphics[width=0.23\linewidth,keepaspectratio,cframe=black 1pt]{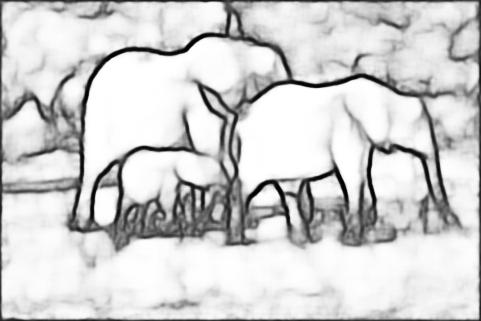} \\
\includegraphics[width=0.23\linewidth,keepaspectratio,cframe=black 1pt]{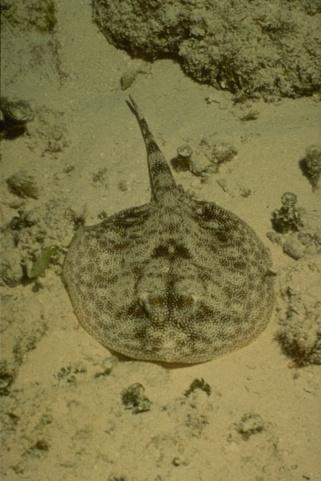} &
\includegraphics[width=0.23\linewidth,keepaspectratio,cframe=black 1pt]{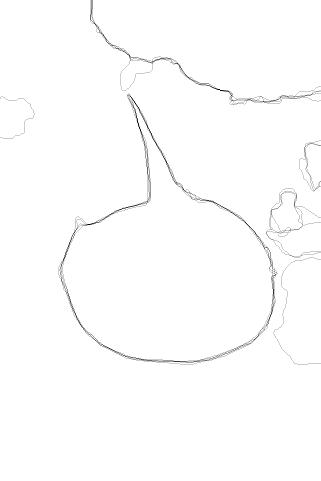} &
\includegraphics[width=0.23\linewidth,keepaspectratio,cframe=black 1pt]{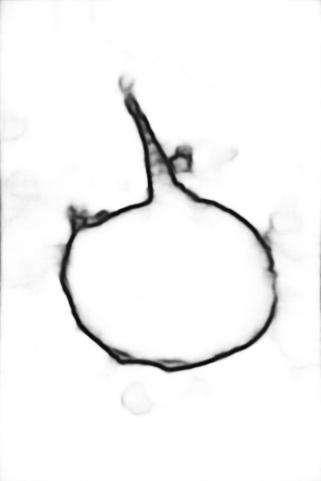} &
\includegraphics[width=0.23\linewidth,keepaspectratio,cframe=black 1pt]{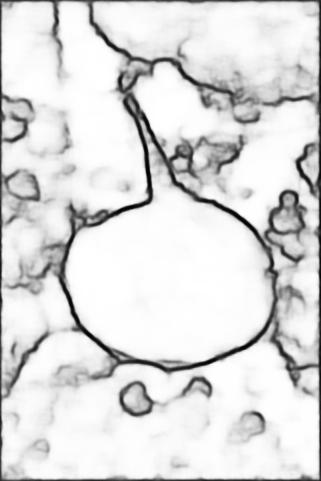} \\
\includegraphics[width=0.23\linewidth,keepaspectratio,cframe=black 1pt]{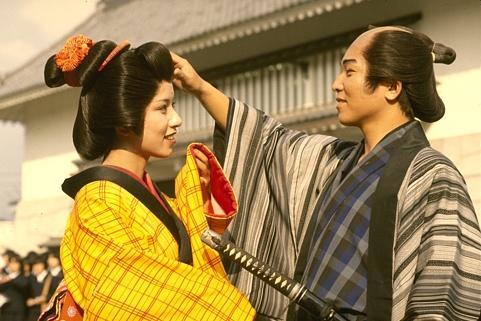} &
\includegraphics[width=0.23\linewidth,keepaspectratio,cframe=black 1pt]{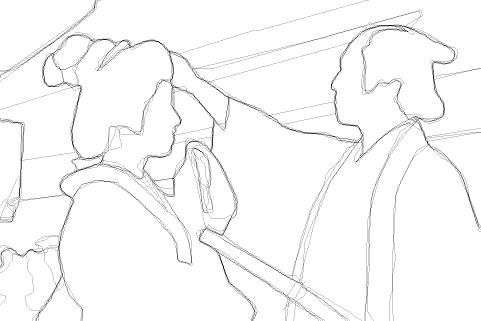} &
\includegraphics[width=0.23\linewidth,keepaspectratio,cframe=black 1pt]{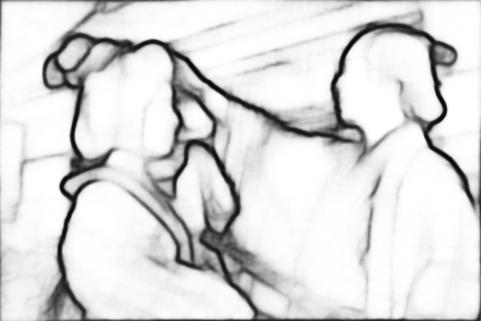} &
\includegraphics[width=0.23\linewidth,keepaspectratio,cframe=black 1pt]{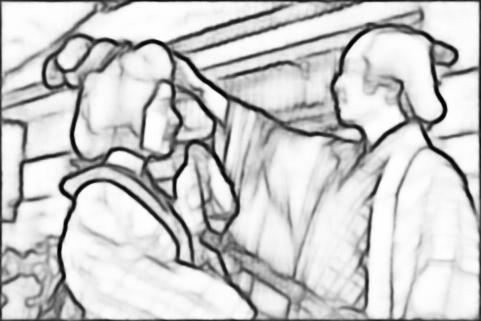} \\
\includegraphics[width=0.23\linewidth,keepaspectratio,cframe=black 1pt]{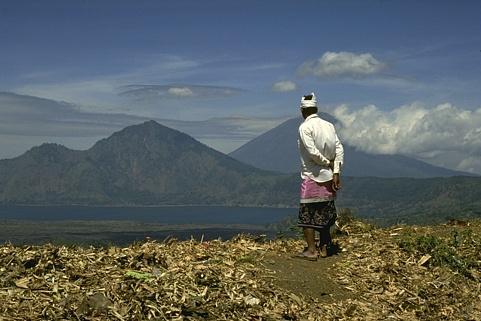} &
\includegraphics[width=0.23\linewidth,keepaspectratio,cframe=black 1pt]{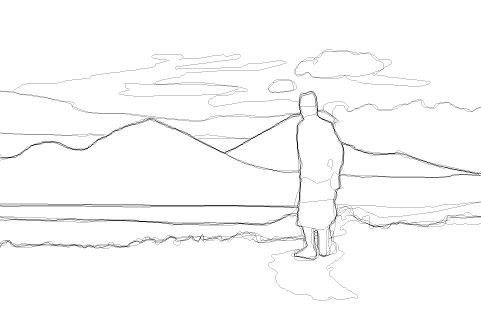} &
\includegraphics[width=0.23\linewidth,keepaspectratio,cframe=black 1pt]{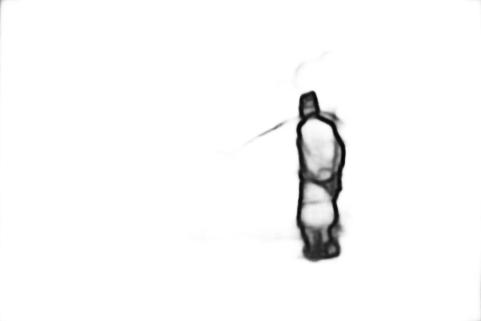} &
\includegraphics[width=0.23\linewidth,keepaspectratio,cframe=black 1pt]{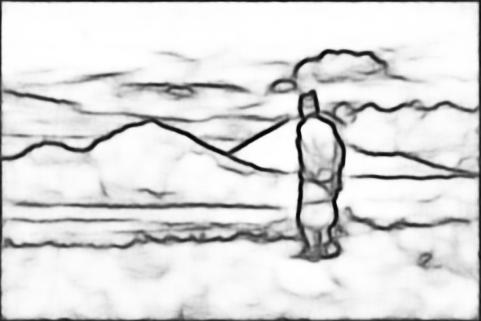} \\
(a) & (b)  & (c) & (d) \\
\end{tabular}
\caption{Example results on BSDS500 test set. In each row from left to right we present (a) input image, (b) ground truth contour, (c) contour detection with pretrained CEDN and (d) contour detection with fine-tuned CEDN.}
\label{fig:bsds_example}
\vspace{-2mm}
\end{figure}
\paragraph{BSDS500 with fine-tuning.}
BSDS500~\cite{Martin_ICCV_2001} is a standard benchmark for contour detection. 
Different from our object-centric goal, this dataset is designed for evaluating natural edge detection that includes not only object contours but also object interior boundaries and background boundaries (examples in Figure~\ref{fig:bsds_example}(b)).
It includes 500 natural images with carefully annotated boundaries collected from multiple users. 
The dataset is divided into three parts: 200 for training, 100 for validation and the rest 200 for test. 
We first examine how well our CEDN model trained on PASCAL VOC can generalize to unseen object categories in this dataset.
Interestingly, as shown in the Figure~\ref{fig:bsds_example}(c), most of wild animal contours, e.g. elephants and fish are accurately detected and meanwhile the background boundaries, e.g. building and mountains are clearly suppressed. 
We further fine-tune our CEDN model on the 200 training images from BSDS500 with a small learning rate ($10^{-5}$) for 100 epochs.
As a result, the boundaries suppressed by pretrained CEDN model (``CEDN-pretrain") re-surface from the scenes. 
Quantitatively, we evaluate both the pretrained and fine-tuned models on the test set in comparisons with previous methods.
\begin{figure}[htbp]
\centering
\includegraphics[width=1.0\linewidth,keepaspectratio]{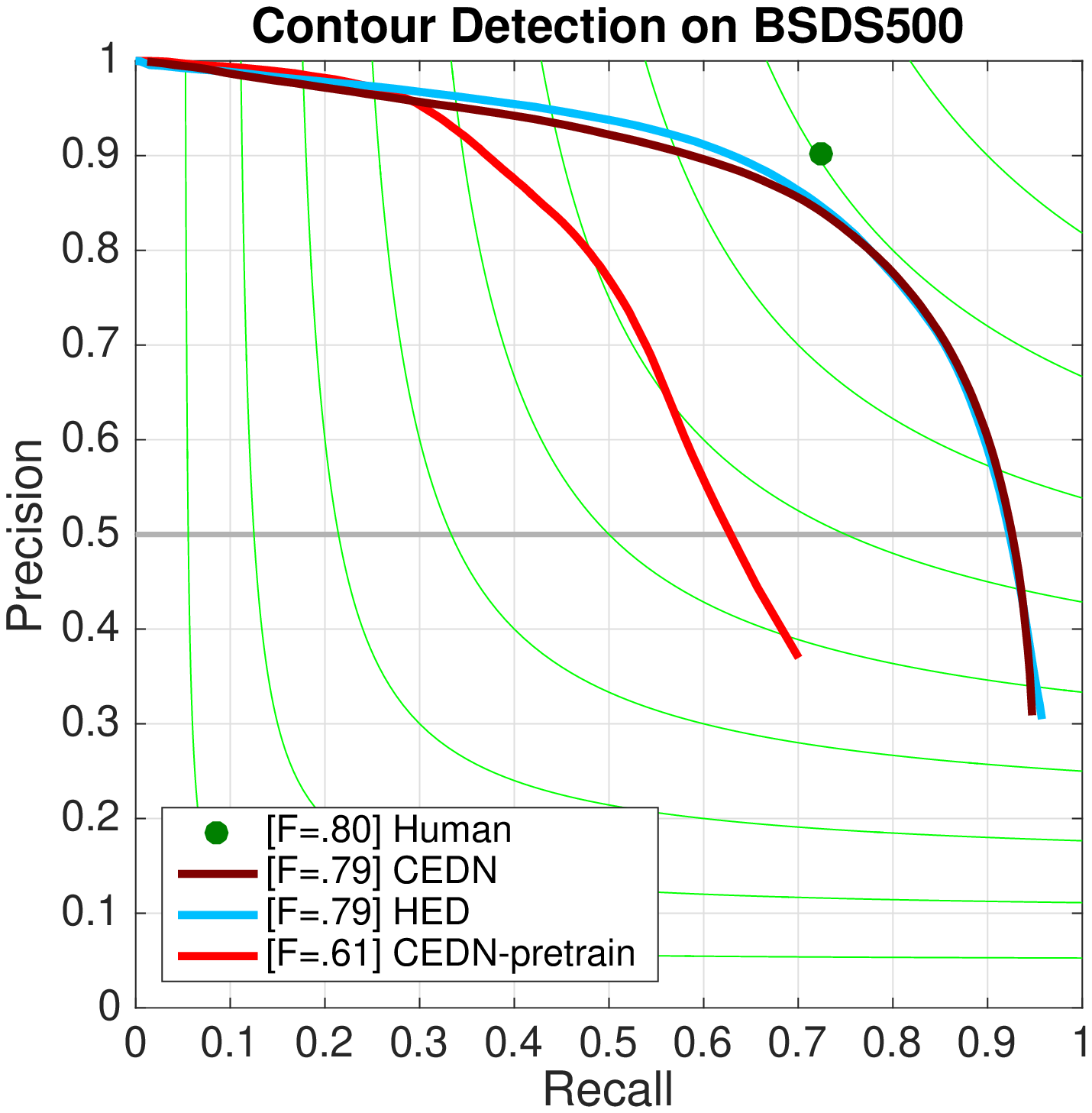}
\caption{PR curve for contour detection on the BSDS500 set set.}
\label{fig:pr_curve_bsds500}
\vspace{-3mm}
\end{figure}
Figure~\ref{fig:pr_curve_bsds500} shows that 1) the pretrained CEDN model yields a high precision but a low recall due to its object-selective nature and 2) the fine-tuned CEDN model achieves comparable performance (F=0.79) with the state-of-the-art method (HED)~\cite{Xie_ICCV_2015}.
Note that our model is not deliberately designed for natural edge detection on BSDS500, and we believe that the techniques used in HED~\cite{Xie_ICCV_2015} such as multiscale fusion, carefully designed upsampling layers and data augmentation could further improve the performance of our model.
A more detailed comparison is listed in Table~\ref{tab:bsds500}.
\begin{table}[!htb]
\centering
\caption{Contour detection results on BSDS500.}
\label{tab:bsds500}
\begin{tabular}{c|c|c|c}
\hline
 & ODS & OIS & AP \\
\Xhline{4\arrayrulewidth}
Human & .80 & .80 & -\\
\hline
SCG~\cite{Arbelaez_CVPR_2014} & .739 & .758 & .773 \\
SE~\cite{Dollar_ICCV_2013} & .746 & .767 & .803 \\
\hline
DeepEdge~\cite{Bertasius_CVPR_2015} & .753 & .772 & .807 \\
DeepContour~\cite{Shen_CVPR_2015} & .756 & .773 & .797 \\
HED~\cite{Xie_ICCV_2015} &  .782 & .804 & .833 \\
HED-new~\footnotemark & \textbf{.788} & \textbf{.808} & \textbf{.840} \\
\hline
CEDN-pretrain & .610 & .635 & .580 \\
CEDN & \textbf{.788} & .804 & .821 \\
\hline
\end{tabular}
\vspace{-3mm}
\end{table}
\addtocounter{footnote}{0}
\footnotetext{This is the latest result from \url{http://vcl.ucsd.edu/hed/}}

\subsection{Object Proposal Generation}
Object proposals are important mid-level representations in computer vision.
Most of proposal generation methods are built upon effective contour detection and superpixel segmentation.
Thus the improvements on contour detection will immediately boost the performance of object proposals.
We choose the MCG algorithm to generate segmented object proposals from our detected contours.
The MCG algorithm is based the classic \emph{gPb} contour detector.
It first computes ultrametric contour maps from multiscale and then aligns them into a single hierarchical segmentation.
To obtain object proposals, a multi-objective optimization is designed to reduce the redundancy of combinatorial grouping of adjacent regions.
The reduced set of grouping candidates are then ranked according to their low-level features as the final segmented object proposals.
Note that the hierarchical segmentation can be obtained directly from single scale, which leads to a fast algorithm of singlescale combinatorial grouping (SCG).
Based on the procedure above, we simply replace \emph{gPb} with our CEDN contour detector to generate proposals.
The multiscale and singlescale versions are referred to as ``CEDNMCG" and ``CEDNSCG", respectively.

We evaluate the quality of object proposals by two measures: Average Recall (AR) and Average Best Overlap (ABO).
Both measures are based on the overlap (Jaccard index or Intersection-over-Union) between a proposal and a ground truth mask.
AR is measured by 1) counting the percentage of objects with their best Jaccard above a certain threshold $T$ and then 2) averaging them within a range of thresholds $T\in[0.5,1.0]$. 
It is established in~\cite{Hosang_PAMI_2015, Pont_Tuset_ICCV_2015} to benchmark the quality of bounding box and segmented object proposals.
ABO is measured by calculating the best proposal's Jaccard for every ground truth object and then 2) averaging them over all the objects.

\begin{figure}[t]
\centering
\includegraphics[width=0.8\linewidth,keepaspectratio]{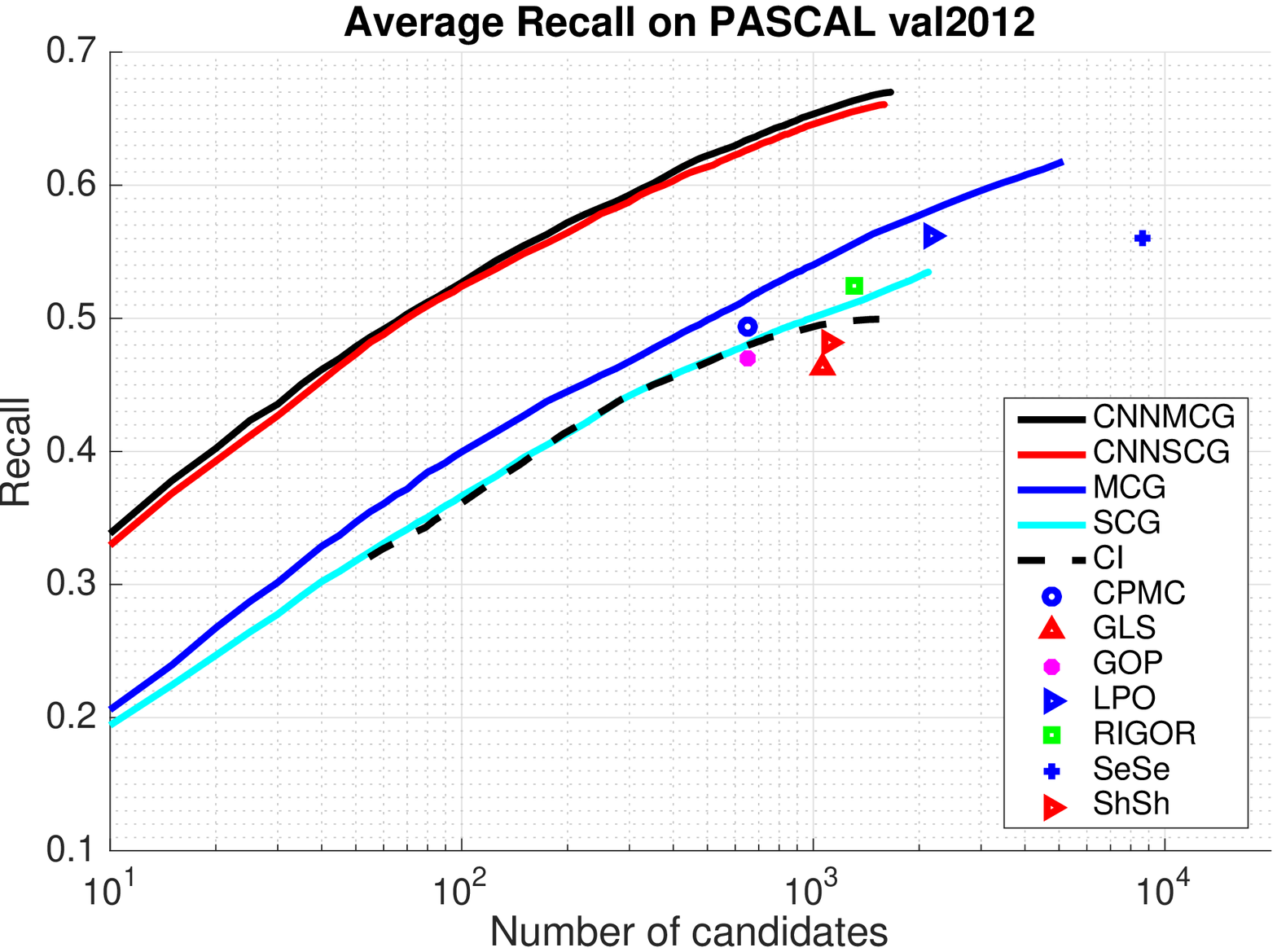}
\includegraphics[width=0.8\linewidth,keepaspectratio]{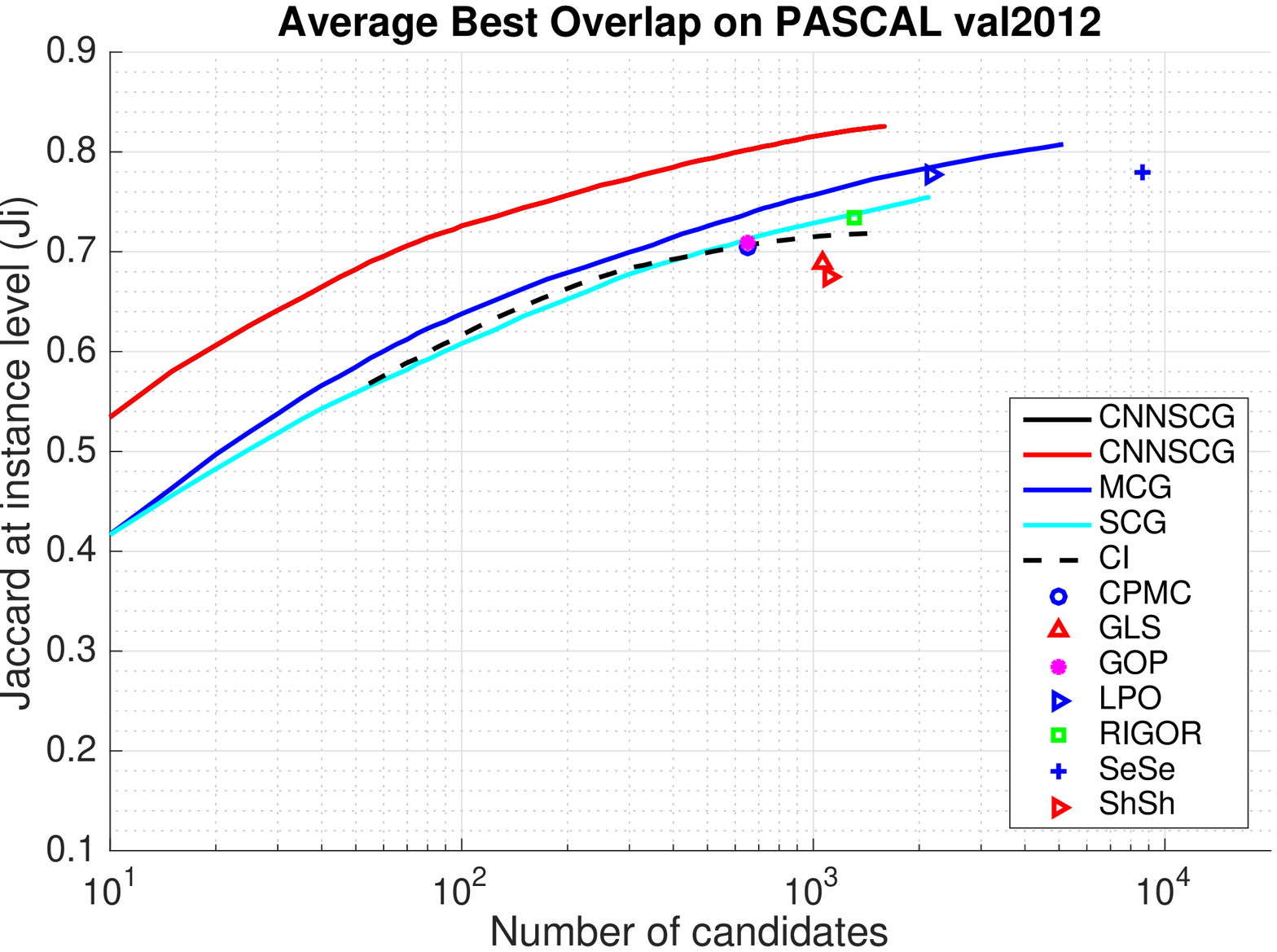}
\caption{Average best overlap and average recall on the PASCAL VOC 2012 validation set.}
\label{fig:abo_ar_pascal2012}
\vspace{-3mm}
\end{figure}
\begin{figure}[htbp]
\centering
\includegraphics[width=0.8\linewidth,keepaspectratio]{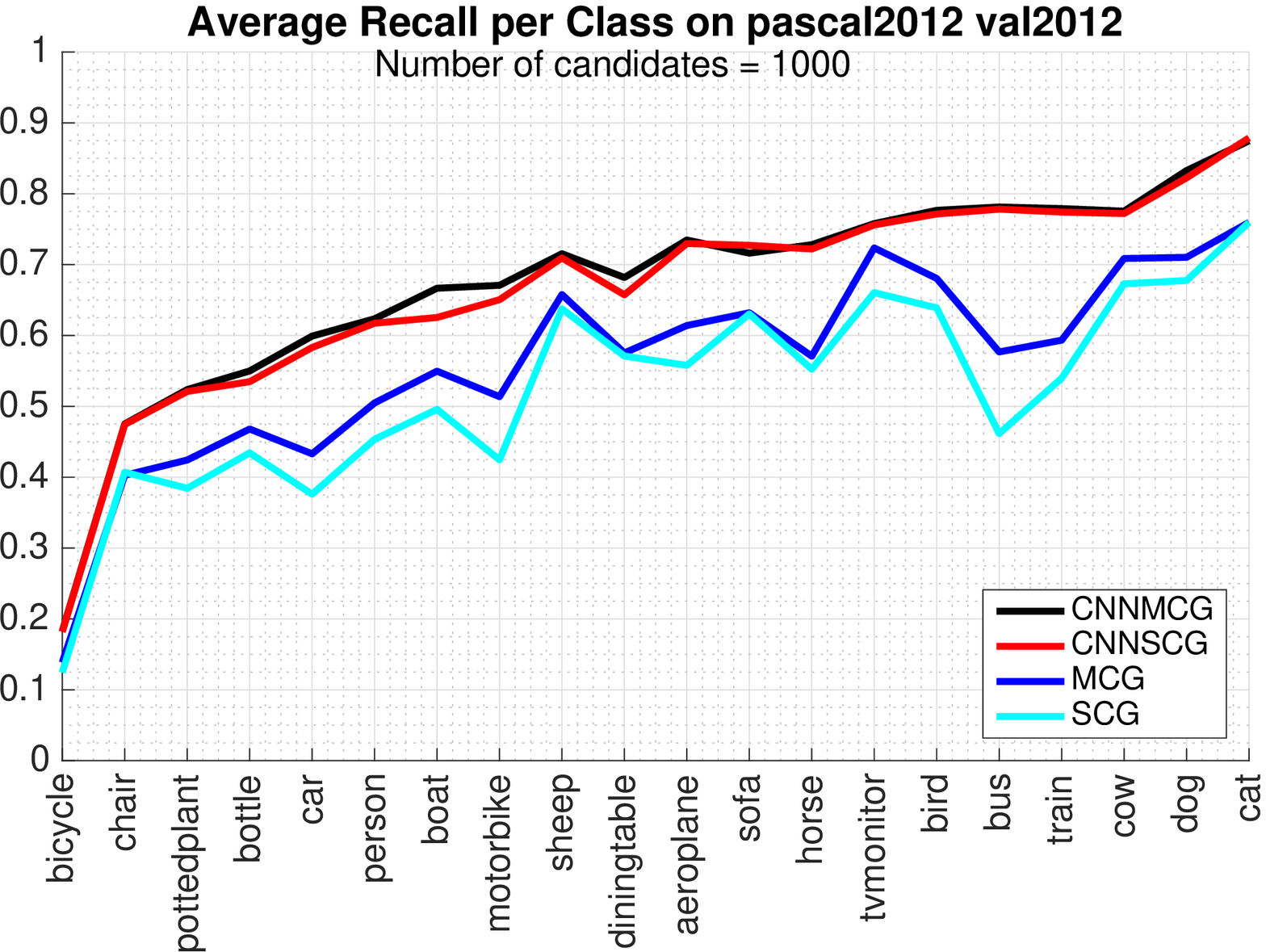}
\caption{Per-class average recall on the PASCAL VOC 2012 validation set.}
\label{fig:ar_class_pascal2012}
\end{figure}

We compare with state-of-the-art algorithms: MCG, SCG, Category Independent object proposals (CI)~\cite{Endres_ECCV_2010}, Constraint Parametric Min Cuts (CPMC)~\cite{Carreira_CVPR_2010}, Global and Local Search (GLS)~\cite{Rantalankila_CVPR_2014}, Geodesic Object Proposals (GOP)~\cite{Krahenbuhl_ECCV_2014}, Learning to Propose Objects (LPO)~\cite{Krahenbuhl_CVPR_2015}, Recycling Inference in Graph Cuts (RIGOR)~\cite{Humayun_CVPR_2014}, Selective Search (SeSe)~\cite{deSande_ICCV_2011} and Shape Sharing (ShSh)~\cite{Kim_ECCV_2012}.
Note that these abbreviated names are inherited from~\cite{Arbelaez_CVPR_2014}.
\begin{figure*}[t]
\centering
\setlength{\tabcolsep}{1pt}
\begin{tabular}{ccccc}
\includegraphics[width=0.18\linewidth,keepaspectratio,cframe=black 1pt]{COCO_val2014_000000001000_im.jpg} &
\includegraphics[width=0.18\linewidth,keepaspectratio,cframe=black 1pt]{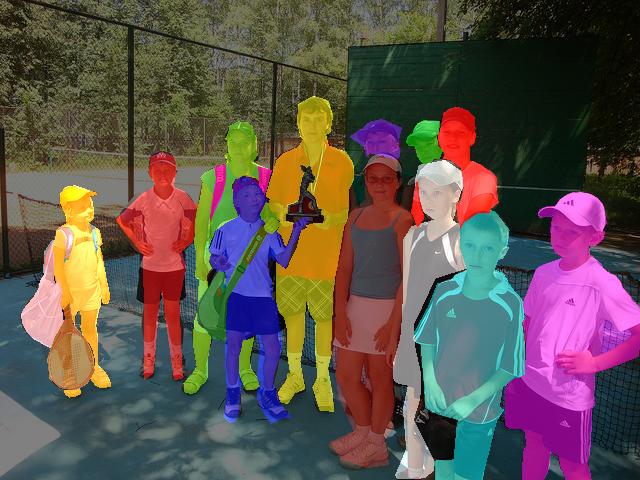} &
\includegraphics[width=0.18\linewidth,keepaspectratio,cframe=black 1pt]{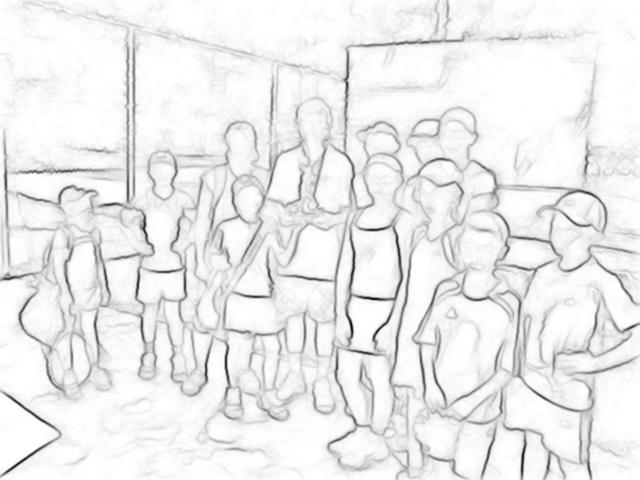} &
\includegraphics[width=0.18\linewidth,keepaspectratio,cframe=black 1pt]{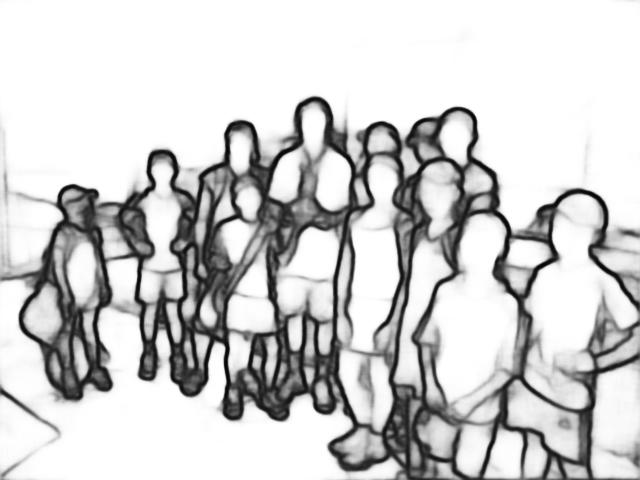} &
\includegraphics[width=0.18\linewidth,keepaspectratio,cframe=black 1pt]{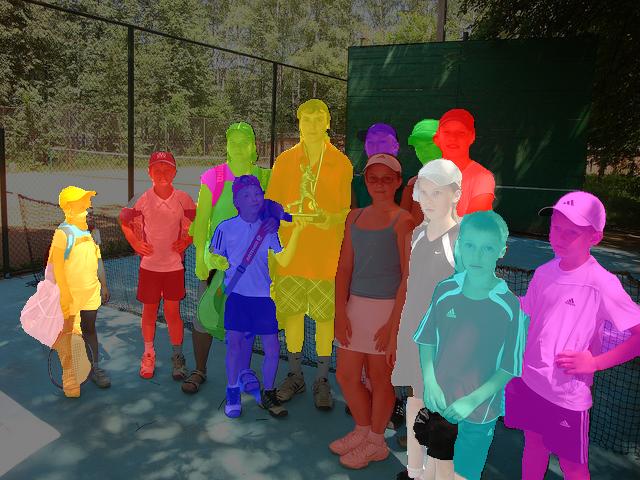} \\
\includegraphics[width=0.18\linewidth,keepaspectratio,cframe=black 1pt]{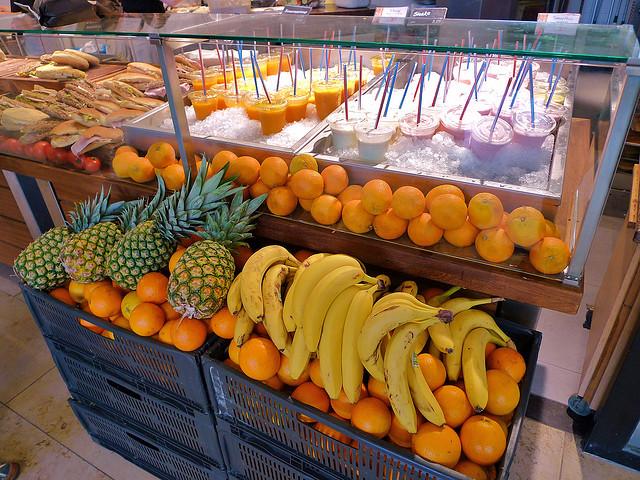} &
\includegraphics[width=0.18\linewidth,keepaspectratio,cframe=black 1pt]{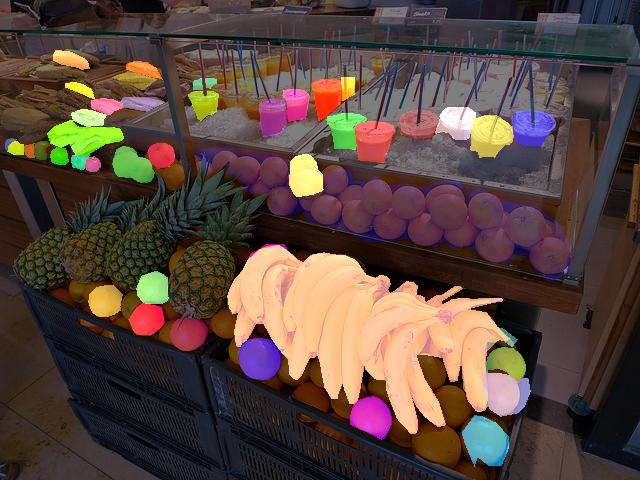} &
\includegraphics[width=0.18\linewidth,keepaspectratio,cframe=black 1pt]{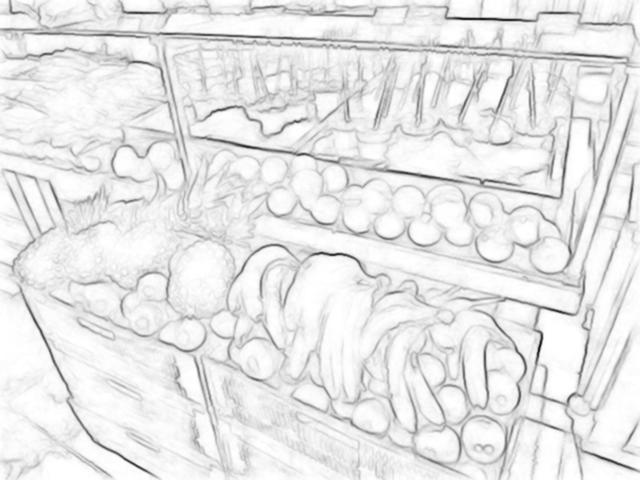} &
\includegraphics[width=0.18\linewidth,keepaspectratio,cframe=black 1pt]{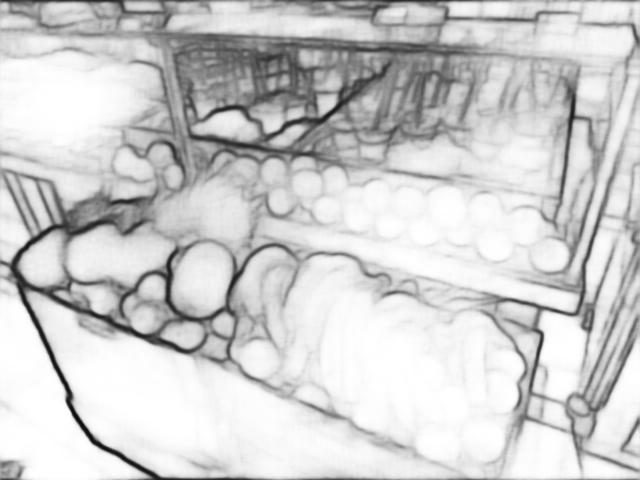} &
\includegraphics[width=0.18\linewidth,keepaspectratio,cframe=black 1pt]{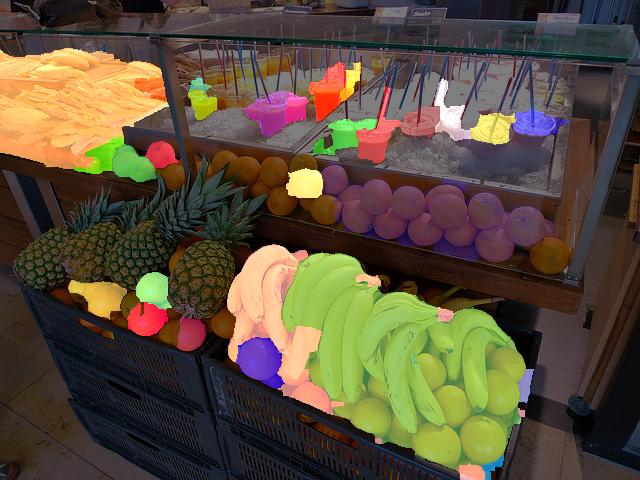} \\
\includegraphics[width=0.18\linewidth,keepaspectratio,cframe=black 1pt]{COCO_val2014_000000004312_im.jpg} &
\includegraphics[width=0.18\linewidth,keepaspectratio,cframe=black 1pt]{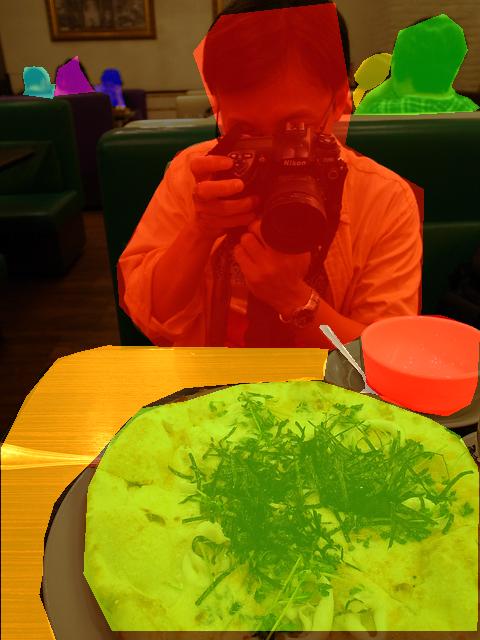} &
\includegraphics[width=0.18\linewidth,keepaspectratio,cframe=black 1pt]{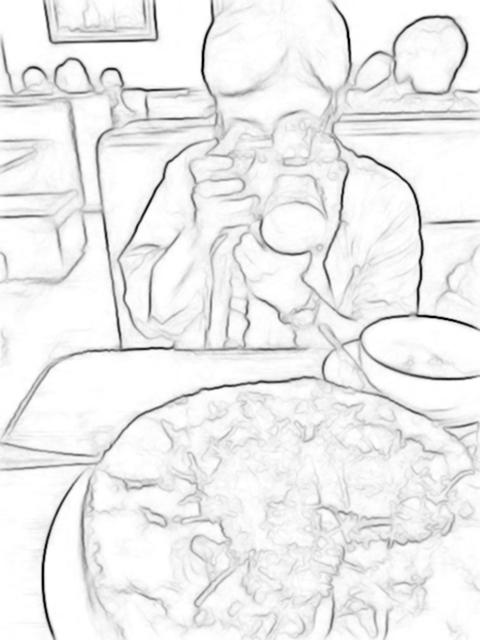} &
\includegraphics[width=0.18\linewidth,keepaspectratio,cframe=black 1pt]{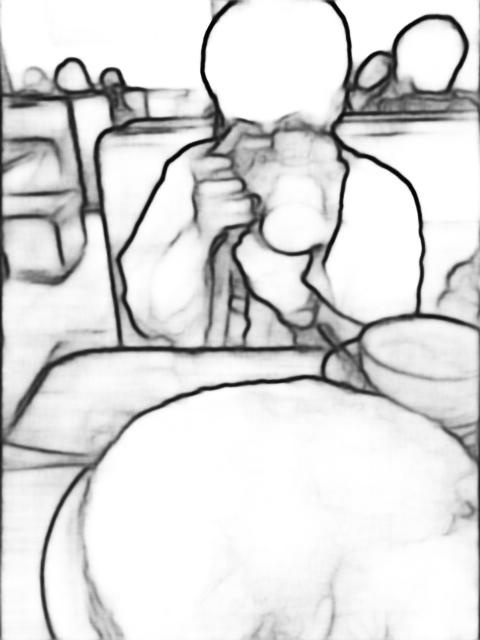} &
\includegraphics[width=0.18\linewidth,keepaspectratio,cframe=black 1pt]{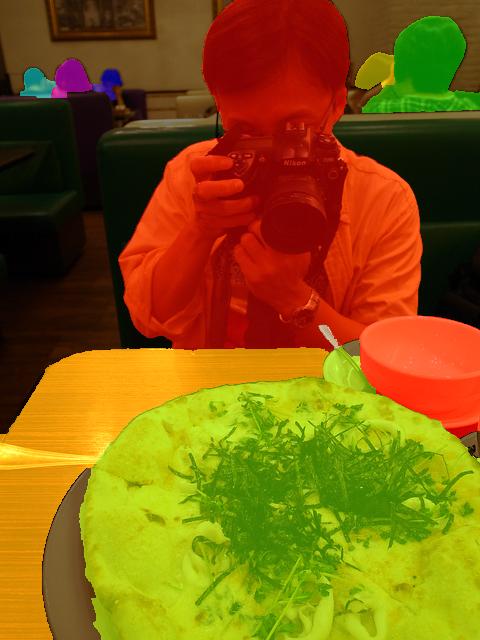} \\
\includegraphics[width=0.18\linewidth,keepaspectratio,cframe=black 1pt]{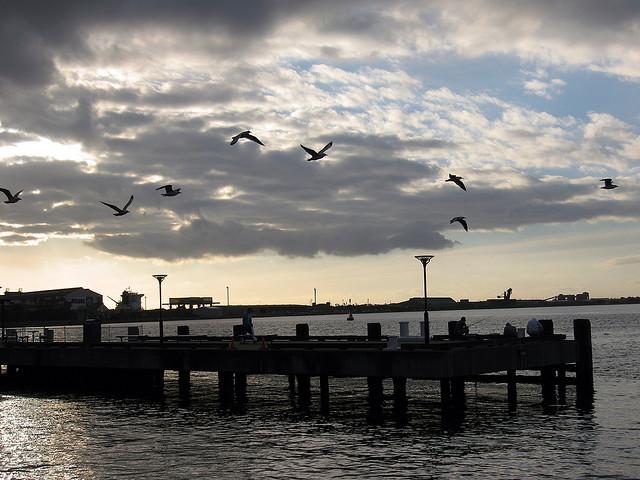} &
\includegraphics[width=0.18\linewidth,keepaspectratio,cframe=black 1pt]{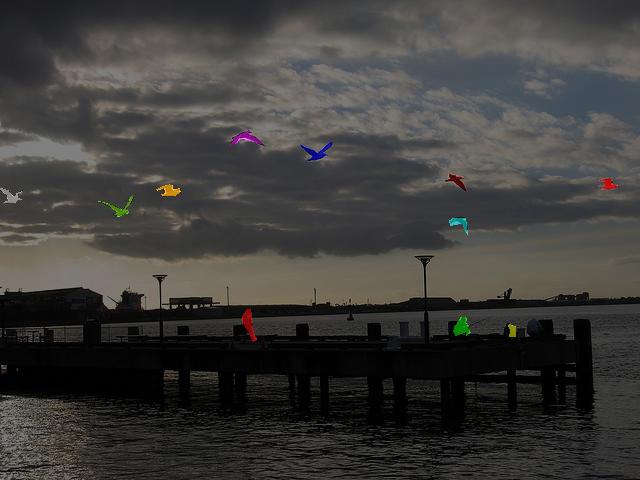} &
\includegraphics[width=0.18\linewidth,keepaspectratio,cframe=black 1pt]{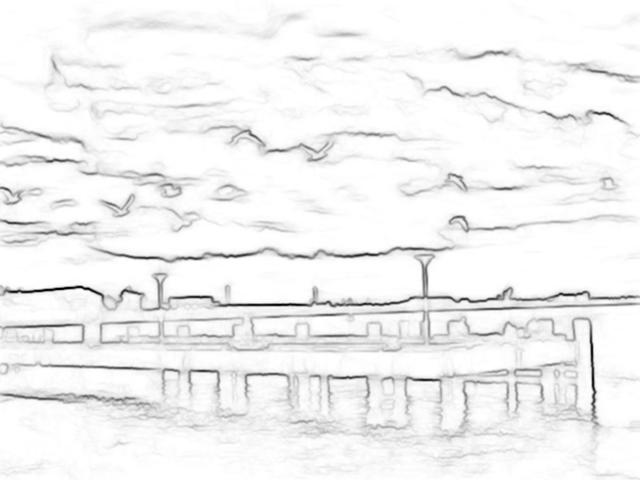} &
\includegraphics[width=0.18\linewidth,keepaspectratio,cframe=black 1pt]{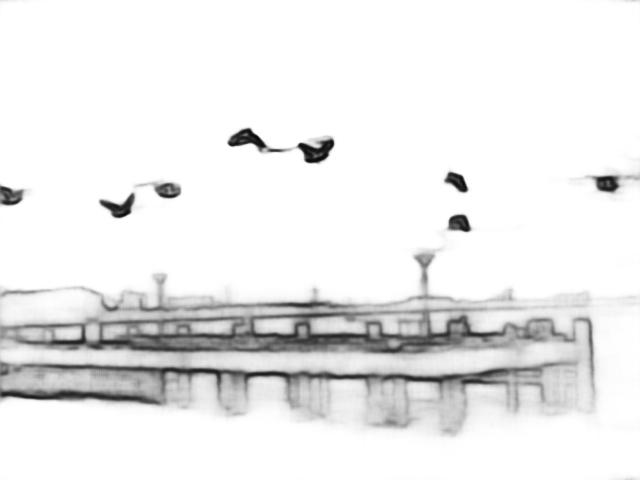} &
\includegraphics[width=0.18\linewidth,keepaspectratio,cframe=black 1pt]{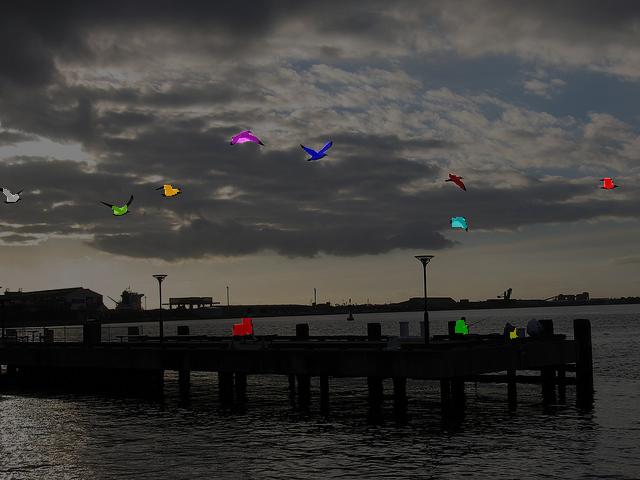} \\
(a) & (b)  & (c) & (d) & (e) \\
\end{tabular}
\caption{Example results on MS COCO val2014. In each row from left to right we present (a) input image, (b) ground truth annotation, (c) edge detection~\cite{Dollar_ICCV_2013}, (d) our object contour detection and (e) our best object proposals.}
\label{fig:coco_example}
\vspace{-2mm}
\end{figure*}
\paragraph{PASCAL val2012.}
Figure~\ref{fig:abo_ar_pascal2012} shows that CEDNMCG achieves 0.67 AR and 0.83 ABO with $\sim1660$ proposals per image, which improves the second best MCG by 8\% in AR and by 3\% in ABO with a third as many proposals.
It takes 0.1 second to compute the CEDN contour map for a PASCAL image on a high-end GPU and 18 seconds to generate proposals with MCG on a standard CPU. 
We notice that the CEDNSCG achieves similar accuracies with CEDNMCG, but it only takes less than 3 seconds to run SCG.
We also plot the per-class ARs in Figure~\ref{fig:ar_class_pascal2012} and find that CEDNMCG and CEDNSCG improves MCG and SCG for all of the 20 classes. 
Notably, the bicycle class has the worst AR and we guess it is likely because of its incomplete annotations.
Some examples of object proposals are demonstrated in Figure~\ref{fig:pascal_example}(d).
\ignore{
\begin{figure*}[h]
\centering
\includegraphics[width=0.3\linewidth,keepaspectratio]{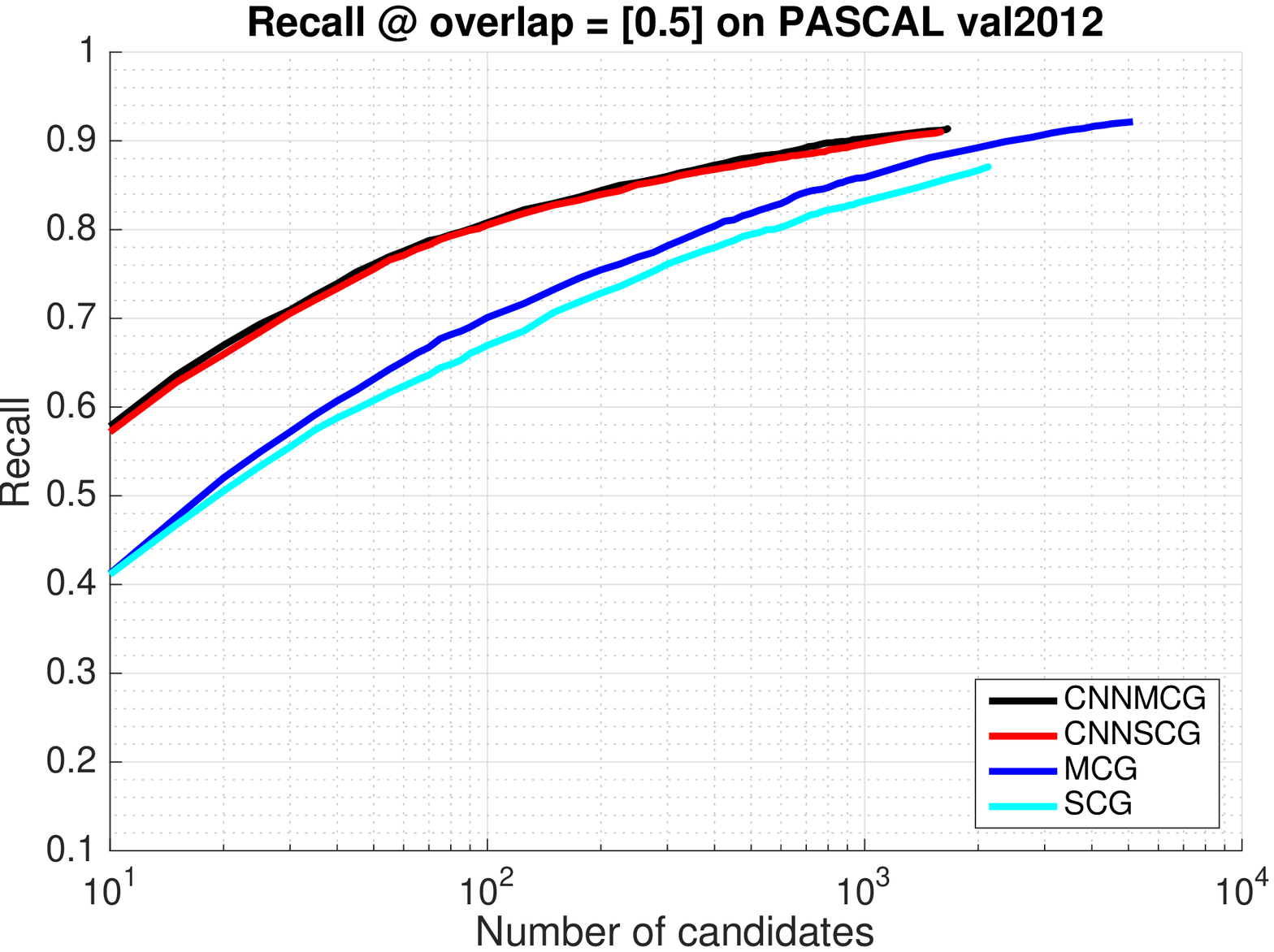}
\includegraphics[width=0.3\linewidth,keepaspectratio]{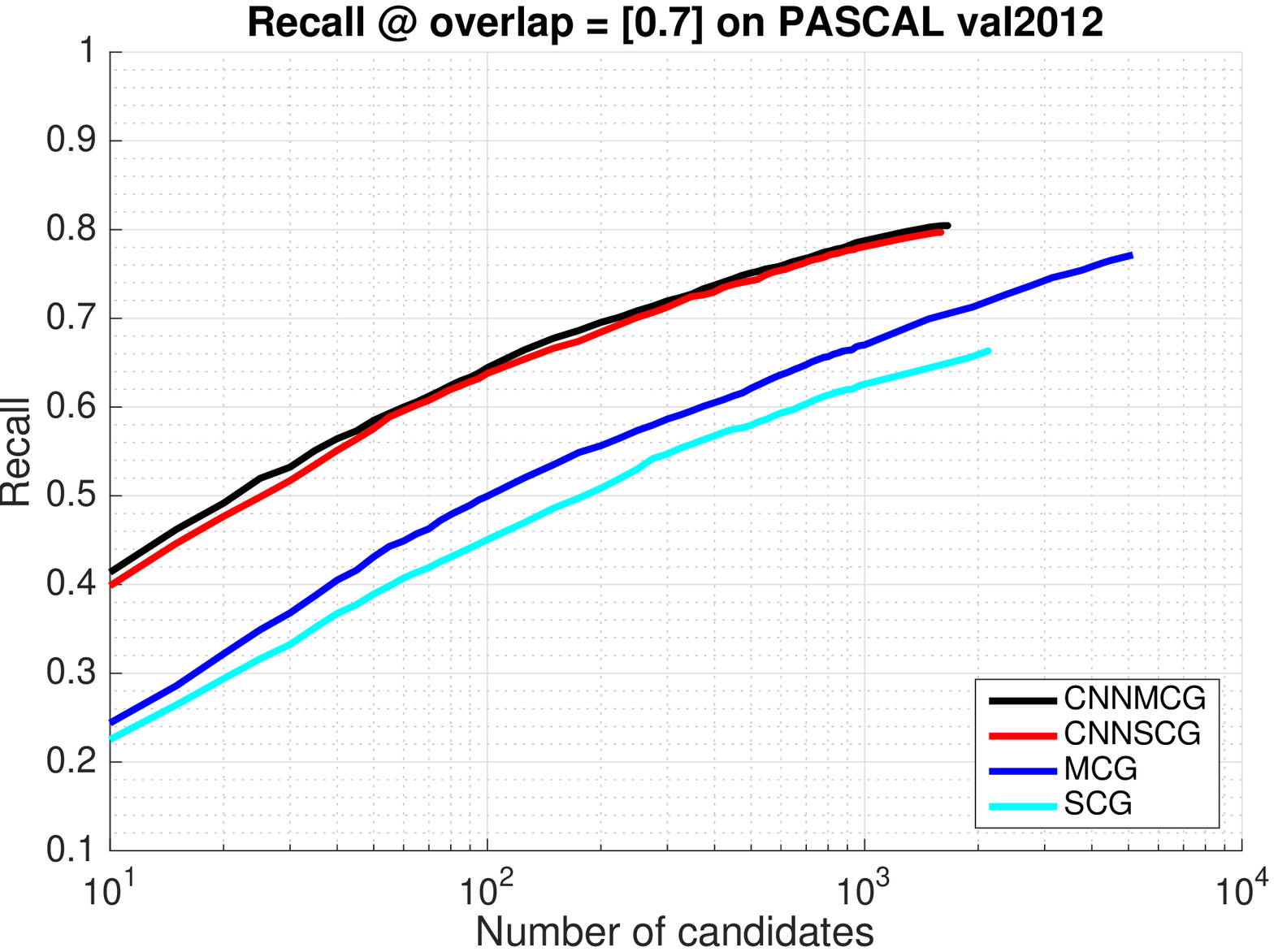}
\includegraphics[width=0.3\linewidth,keepaspectratio]{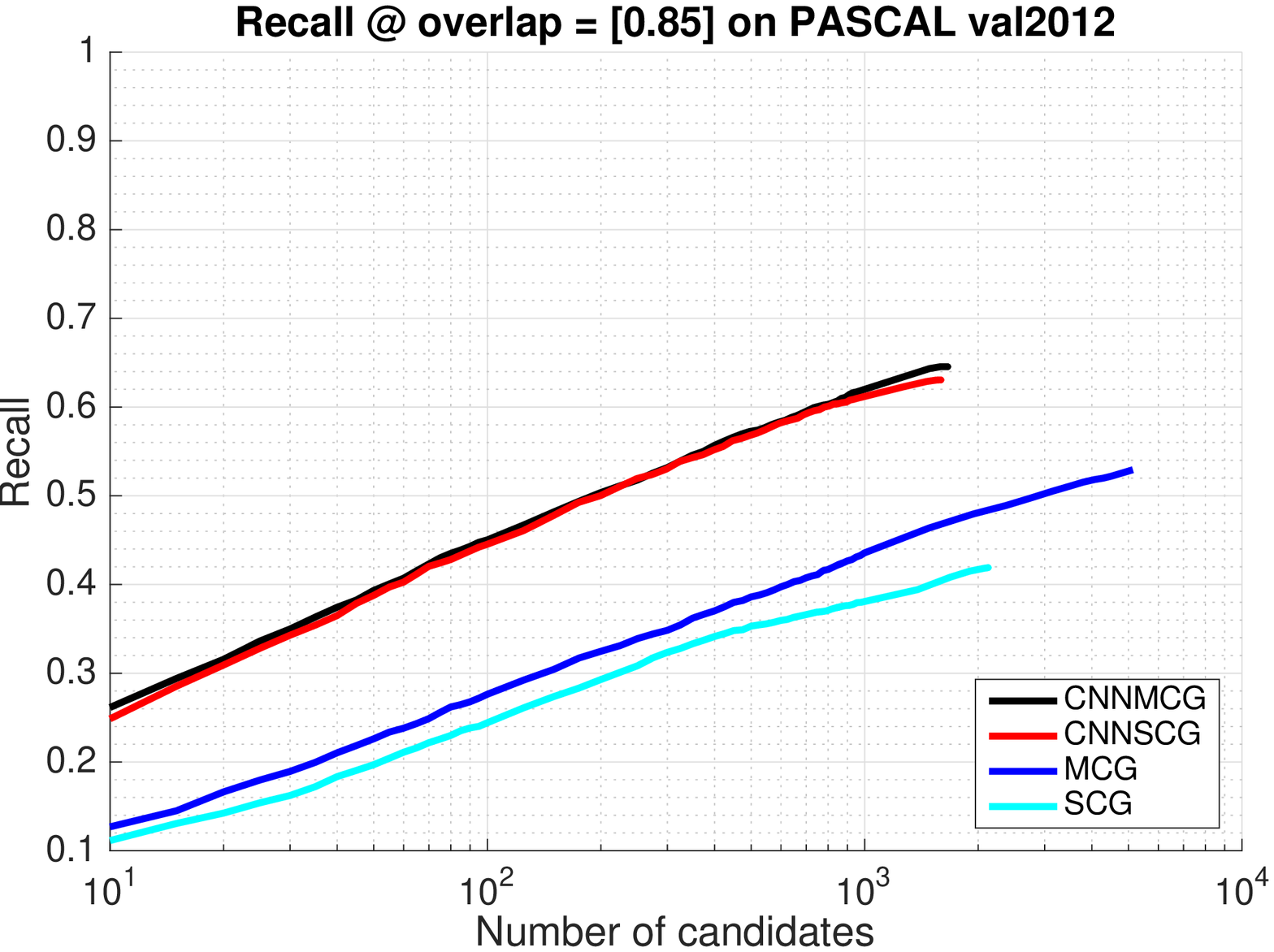}
\caption{Object recalls at overlap rates = [0.5,0.7,0.85] on the PASCAL VOC 2012 validation set.}
\label{fig:recalls_voc2012}
\end{figure*}
}
\paragraph{MS COCO val2014.}
\begin{figure*}[t]
\centering
\includegraphics[width=0.4\linewidth,keepaspectratio]{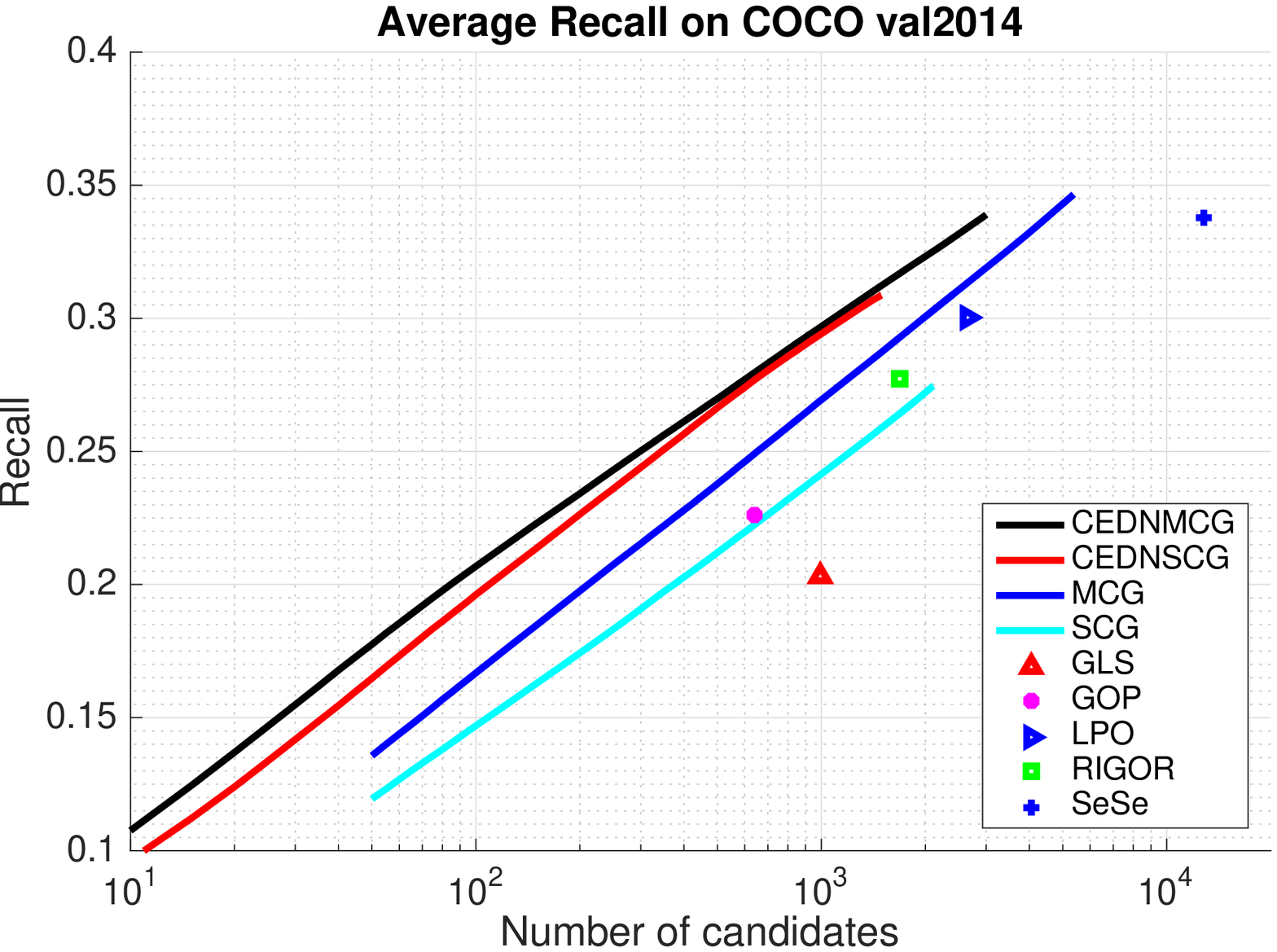}
\includegraphics[width=0.4\linewidth,keepaspectratio]{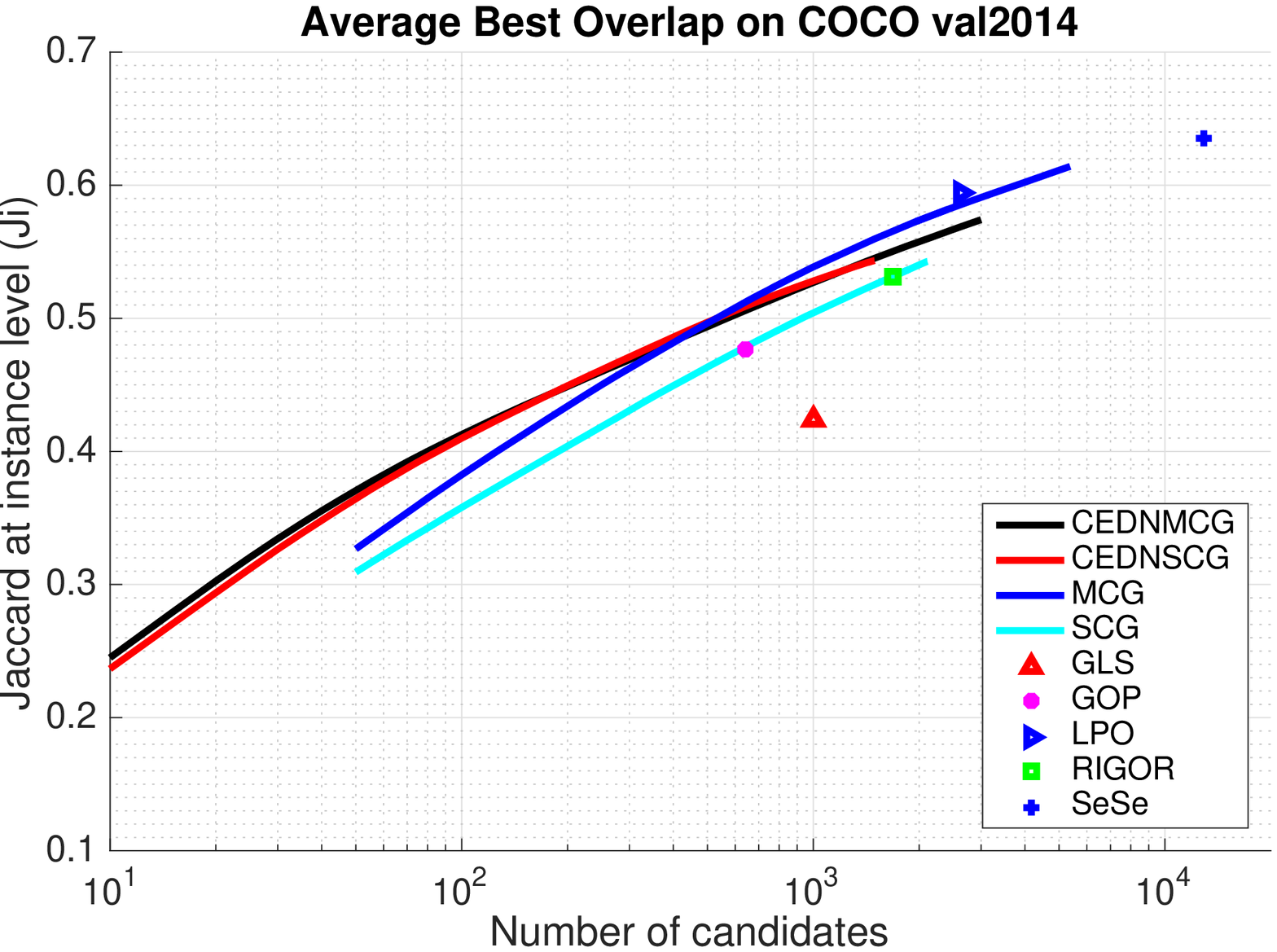}
\caption{Average best overlap and average recall on the MS COCO 2014 validation set.}
\label{fig:abo_ar_coco}
\end{figure*}
\begin{figure*}[htbp]
\centering
\includegraphics[width=0.85\linewidth,keepaspectratio]{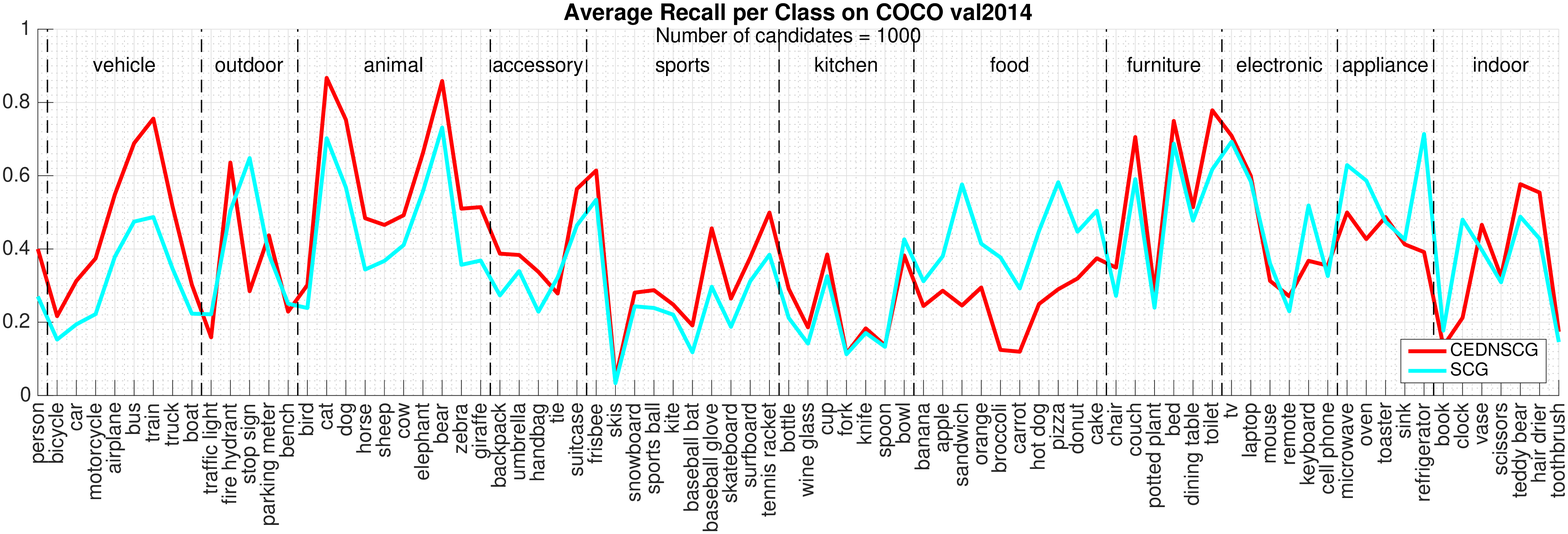}
\caption{Average recall per class on the MS COCO 2014 validation set.}
\label{fig:ar_class_coco}
\end{figure*}
We present results in the MS COCO 2014 validation set, shortly ``COCO val2014" that includes 40504 images annotated by polygons from 80 object classes.
This dataset is more challenging due to its large variations of object categories, contexts and scales.
Compared to PASCAL VOC, there are 60 unseen object classes for our CEDN contour detector.
Note that we did not train CEDN on MS COCO.
We report the AR and ABO results in Figure~\ref{fig:abo_ar_coco}.
It turns out that the CEDNMCG achieves a competitive AR to MCG with a slightly lower recall from fewer proposals, but a weaker ABO than LPO, MCG and SeSe.
Taking a closer look at the results, we find that our CEDNMCG algorithm can still perform well on \emph{known} objects (first and third examples in Figure~\ref{fig:coco_example}) but less effectively on certain \emph{unknown} object classes, such as food (second example in Figure~\ref{fig:coco_example}). 
It is likely because those novel classes, although seen in our training set (PASCAL VOC), are actually annotated as background.
For example, there is a ``dining table" class but no ``food" class in the PASCAL VOC dataset.
Quantitatively, we present per-class ARs in Figure~\ref{fig:ar_class_coco} and have following observations:
\begin{itemize}
\item CEDN obtains good results on those classes that share common super-categories with PASCAL classes, such as ``vehicle", ``animal" and ``furniture".
\item CEDN fails to detect the objects labeled as ``background" in the PASCAL VOC training set, such as ``food" and ``applicance".
\item CEDN works well on unseen classes that are not prevalent in the PASCAL VOC training set, such as ``sports".
\end{itemize}
These observations urge training on COCO, but we also observe that the polygon annotations in MS COCO are less reliable than the ones in PASCAL VOC (third example in Figure~\ref{fig:coco_example}(b)).
We will need more sophisticated methods for refining the COCO annotations.

\section{Conclusion and Future Work}
We have developed an object-centric contour detection method using a simple yet efficient fully convolutional encoder-decoder network. 
Concerned with the imperfect contour annotations from polygons, we have developed a refinement method based on dense CRF so that the proposed network has been trained in an end-to-end manner.
As a result, the trained model yielded high precision on PASCAL VOC and BSDS500, and has achieved comparable performance with the state-of-the-art on BSDS500 after fine-tuning.
We have combined the proposed contour detector with multiscale combinatorial grouping algorithm for generating segmented object proposals, which significantly advances the state-of-the-art on PASCAL VOC.
We also found that the proposed model generalizes well to unseen object classes from the known super-categories and demonstrated competitive performance on MS COCO without re-training the network.

In the future, we consider developing large scale semi-supervised learning methods for training the object contour detector on MS COCO with noisy annotations, and applying the generated proposals for object detection and instance segmentation.

{\small
\bibliographystyle{ieee}
\bibliography{cvpr16_contour}

\begin{thebibliography}{10}\itemsep=-1pt

\bibitem{Alexe_PAMI_2012}
B.~Alexe, T.~Deselaers, and V.~Ferrari.
\newblock Measuring the objectness of image windows.
\newblock {\em PAMI}, 34:2189--2202, 2012.

\bibitem{Amer_IJCV_2015}
M.~R. Amer, S.~Yousefi, R.~Raich, and S.~Todorovic.
\newblock Monocular extraction of 2.1 {D} sketch using constrained convex
  optimization.
\newblock {\em IJCV}, 112(1):23--42, 2015.

\bibitem{Arbelaez_PAMI_2011}
P.~Arbel{\'a}ez, M.~Maire, C.~Fowlkes, and J.~Malik.
\newblock Contour detection and hierarchical image segmentation.
\newblock {\em PAMI}, 33(5):898--916, 2011.

\bibitem{Arbelaez_CVPR_2014}
P.~Arbel{\'a}ez, J.~Pont-Tuset, J.~Barron, F.~Marques, and J.~Malik.
\newblock Multiscale combinatorial grouping.
\newblock In {\em CVPR}, 2014.

\bibitem{Ba_ICLR_2015}
J.~Ba and D.~Kingma.
\newblock Adam: A method for stochastic optimization.
\newblock In {\em ICLR}, 2015.

\bibitem{Bertasius_CVPR_2015}
G.~Bertasius, J.~Shi, and L.~Torresani.
\newblock Deepedge: A multi-scale bifurcated deep network for top-down contour
  detection.
\newblock In {\em CVPR}, 2015.

\bibitem{Boykov_ICCV_2001}
Y.~Y. Boykov and M.-P. Jolly.
\newblock Interactive graph cuts for optimal boundary \& region segmentation of
  objects in n-d images.
\newblock In {\em ICCV}, 2001.

\bibitem{Canny_PAMI_1986}
J.~Canny.
\newblock A computational approach to edge detection.
\newblock {\em PAMI}, (6):679--698, 1986.

\bibitem{Carreira_CVPR_2010}
J.~Carreira and C.~Sminchisescu.
\newblock Constrained parametric min-cuts for automatic object segmentation.
\newblock In {\em CVPR}, 2010.

\bibitem{Chen_arXiv_2014}
L.-C. Chen, G.~Papandreou, I.~Kokkinos, K.~Murphy, and A.~L. Yuille.
\newblock Semantic image segmentation with deep convolutional nets and fully
  connected crfs.
\newblock {\em arXiv preprint arXiv:1412.7062}, 2014.

\bibitem{Cheng_CVPR_2014}
M.-M. Cheng, Z.~Zhang, W.-Y. Lin, and P.~Torr.
\newblock {BING}: Binarized normed gradients for objectness estimation at
  300fps.
\newblock In {\em CVPR}, 2014.

\bibitem{Dollar_ICCV_2013}
P.~Doll{\'a}r and C.~Zitnick.
\newblock Structured forests for fast edge detection.
\newblock In {\em ICCV}, 2013.

\bibitem{Endres_ECCV_2010}
I.~Endres and D.~Hoiem.
\newblock Category independent object proposals.
\newblock In {\em ECCV}, 2010.

\bibitem{PASCAL_VOC_2010}
M.~Everingham, L.~J.~V. Gool, C.~K.~I. Williams, J.~M. Winn, and A.~Zisserman.
\newblock The {Pascal} visual object classes ({VOC}) challenge.
\newblock {\em IJCV}, 88(2):303--338, 2010.

\bibitem{Ferrari_PAMI_2008}
V.~Ferrari, L.~Fevrier, F.~Jurie, and C.~Schmid.
\newblock Groups of adjacent contour segments for object detection.
\newblock {\em PAMI}, 30(1):36--51, 2008.

\bibitem{Forsyth_CV_2003}
D.~A. Forsyth and J.~Ponce.
\newblock {\em Computer Vision: A Modern Approach}.
\newblock Pearson Education, 2003.

\bibitem{Girshick_ICCV_2015}
R.~Girshick.
\newblock Fast {R-CNN}.
\newblock In {\em ICCV}, 2015.

\bibitem{Girshick_CVPR_2014}
R.~Girshick, J.~Donahue, T.~Darrell, and J.~Malik.
\newblock Rich feature hierarchies for accurate object detection and semantic
  segmentation.
\newblock In {\em CVPR}, 2014.

\bibitem{Hariharan_ICCV_2011}
B.~Hariharan, P.~Arbel{\'a}ez, L.~Bourdev, S.~Maji, and J.~Malik.
\newblock Semantic contours from inverse detectors.
\newblock In {\em ICCV}, 2011.

\bibitem{Hoiem_ICCV_2007}
D.~Hoiem, A.~N. Stein, A.~Efros, and M.~Hebert.
\newblock Recovering occlusion boundaries from a single image.
\newblock In {\em ICCV}, 2007.

\bibitem{Hosang_PAMI_2015}
J.~Hosang, R.~Benenson, P.~Doll{\'a}r, and B.~Schiele.
\newblock What makes for effective detection proposals?
\newblock {\em PAMI}, 2015.

\bibitem{Humayun_CVPR_2014}
A.~Humayun, F.~Li, and J.~M. Rehg.
\newblock {RIGOR}: Reusing inference in graph cuts for generating object
  regions.
\newblock In {\em CVPR}, 2014.

\bibitem{Jia_Caffe_2014}
Y.~Jia, E.~Shelhamer, J.~Donahue, S.~Karayev, J.~Long, R.~Girshick,
  S.~Guadarrama, and T.~Darrell.
\newblock Caffe: Convolutional architecture for fast feature embedding.
\newblock {\em arXiv preprint arXiv:1408.5093}, 2014.

\bibitem{Kim_ECCV_2012}
J.~Kim and K.~Grauman.
\newblock Shape sharing for object segmentation.
\newblock In {\em ECCV}, 2012.

\bibitem{Kivinen_AISTATS_2014}
J.~J. Kivinen, C.~K. Williams, and N.~Heess.
\newblock Visual boundary prediction: A deep neural prediction network and
  quality dissection.
\newblock In {\em AISTATS}, 2014.

\bibitem{Krahenbuhl_NIPS_2011}
P.~Kr{\"a}henb{\"u}hl and V.~Koltun.
\newblock Efficient inference in fully connected {CRF}s with gaussian edge
  potentials.
\newblock In {\em NIPS}, 2011.

\bibitem{Krahenbuhl_ECCV_2014}
P.~Kr{\"a}henb{\"u}hl and V.~Koltun.
\newblock Geodesic object proposals.
\newblock In {\em ECCV}, 2014.

\bibitem{Krahenbuhl_CVPR_2015}
P.~Kr{\"a}henb{\"u}hl and V.~Koltun.
\newblock Learning to propose objects.
\newblock In {\em CVPR}, 2015.

\bibitem{Krizhevsky_NIPS_2012}
A.~Krizhevsky, I.~Sutskever, and G.~E. Hinton.
\newblock Imagenet classification with deep convolutional neural networks.
\newblock In {\em NIPS}, 2012.

\bibitem{Lim_CVPR_2013}
J.~J. Lim, C.~L. Zitnick, and P.~Doll{\'a}r.
\newblock Sketch tokens: A learned mid-level representation for contour and
  object detection.
\newblock In {\em CVPR}, 2013.

\bibitem{Lin_COCO_2014}
T.-Y. Lin, M.~Maire, S.~Belongie, J.~Hays, P.~Perona, D.~Ramanan,
  P.~Doll{\'a}r, and C.~L. Zitnick.
\newblock Microsoft {COCO}: Common objects in context.
\newblock In {\em ECCV}, 2014.

\bibitem{Sifei_CVPR_2015}
S.~Liu, J.~Yang, C.~Huang, and M.-H. Yang.
\newblock Multi-objective convolutional learning for face labeling.
\newblock In {\em CVPR}, 2015.

\bibitem{Liu_ICCV_2015}
Z.~Liu, X.~Li, P.~Luo, C.~C. Loy, and X.~Tang.
\newblock Semantic image segmentation via deep parsing network.
\newblock In {\em ICCV}, 2015.

\bibitem{Long_CVPR_2015}
J.~Long, E.~Shelhamer, and T.~Darrell.
\newblock Fully convolutional networks for semantic segmentation.
\newblock In {\em CVPR}, 2015.

\bibitem{Malik_IJCV_2001}
J.~Malik, S.~Belongie, T.~Leung, and J.~Shi.
\newblock Contour and texture analysis for image segmentation.
\newblock {\em IJCV}, 2001.

\bibitem{Martin_ICCV_2001}
D.~Martin, C.~Fowlkes, D.~Tal, and J.~Malik.
\newblock A database of human segmented natural images and its application to
  evaluating segmentation algorithms and measuring ecological statistics.
\newblock In {\em ICCV}, 2001.

\bibitem{Martin_PAMI_2004}
D.~R. Martin, C.~C. Fowlkes, and J.~Malik.
\newblock Learning to detect natural image boundaries using local brightness,
  color, and texture cues.
\newblock {\em PAMI}, 26(5):530--549, 2004.

\bibitem{Noh_ICCV_2015}
H.~Noh, S.~Hong, and B.~Han.
\newblock Learning deconvolution network for semantic segmentation.
\newblock In {\em ICCV}, 2015.

\bibitem{Pont_Tuset_ICCV_2015}
J.~Pont-Tuset and L.~J.~V. Gool.
\newblock Boosting object proposals: From {Pascal} to {COCO}.
\newblock In {\em ICCV}, 2015.

\bibitem{Rantalankila_CVPR_2014}
P.~Rantalankila, J.~Kannala, and E.~Rahtu.
\newblock Generating object segmentation proposals using global and local
  search.
\newblock In {\em CVPR}, 2014.

\bibitem{Rother_SIGGRAPH_2004}
C.~Rother, V.~Kolmogorov, and A.~Blake.
\newblock Grabcut -interactive foreground extraction using iterated graph cuts.
\newblock {\em ACM Transactions on Graphics (SIGGRAPH)}, 2004.

\bibitem{Russell_IJCV_2008}
B.~C. Russell, A.~Torralba, K.~P. Murphy, and W.~T. Freeman.
\newblock {LabelMe}: a database and web-based tool for image annotation.
\newblock {\em IJCV}, 2008.

\bibitem{Shen_CVPR_2015}
W.~Shen, X.~Wang, Y.~Wang, X.~Bai, and Z.~Zhang.
\newblock Deepcontour: A deep convolutional feature learned by positive-sharing
  loss for contour detection.
\newblock In {\em CVPR}, 2015.

\bibitem{Silberman_ECCV_2012}
N.~Silberman, P.~Kohli, D.~Hoiem, and R.~Fergus.
\newblock Indoor segmentation and support inference from rgbd images.
\newblock In {\em ECCV}, 2012.

\bibitem{Simonyan_ICLR_2015}
K.~Simonyan and A.~Zisserman.
\newblock Very deep convolutional networks for large-scale image recognition.
\newblock In {\em ICLR}, 2015.

\bibitem{deSande_ICCV_2011}
K.~E.~A. van~de Sande, J.~R.~R. Uijlingsy, T.~Gevers, and A.~W.~M. Smeulders.
\newblock Segmentation as selective search for object recognition.
\newblock In {\em ICCV}, 2011.

\bibitem{Xie_ICCV_2015}
S.~Xie and Z.~Tu.
\newblock Holistically-nested edge detection.
\newblock In {\em ICCV}, 2015.

\bibitem{CRFasRNN_ICCV_2015}
S.~Zheng, S.~Jayasumana, B.~Romera-Paredes, V.~Vineet, Z.~Su, D.~Du, C.~Huang,
  and P.~Torr.
\newblock Conditional random fields as recurrent neural networks.
\newblock In {\em ICCV}, 2015.

\bibitem{Zitnick_ECCV_2014}
C.~L. Zitnick and P.~Doll{\'a}r.
\newblock Edge boxes: Locating object proposals from edge.
\newblock In {\em ECCV}, 2014.

\end{thebibliography}
}

\section*{Appendix}

\begin{figure}[t]
\centering
\includegraphics[width=0.8\linewidth,keepaspectratio]{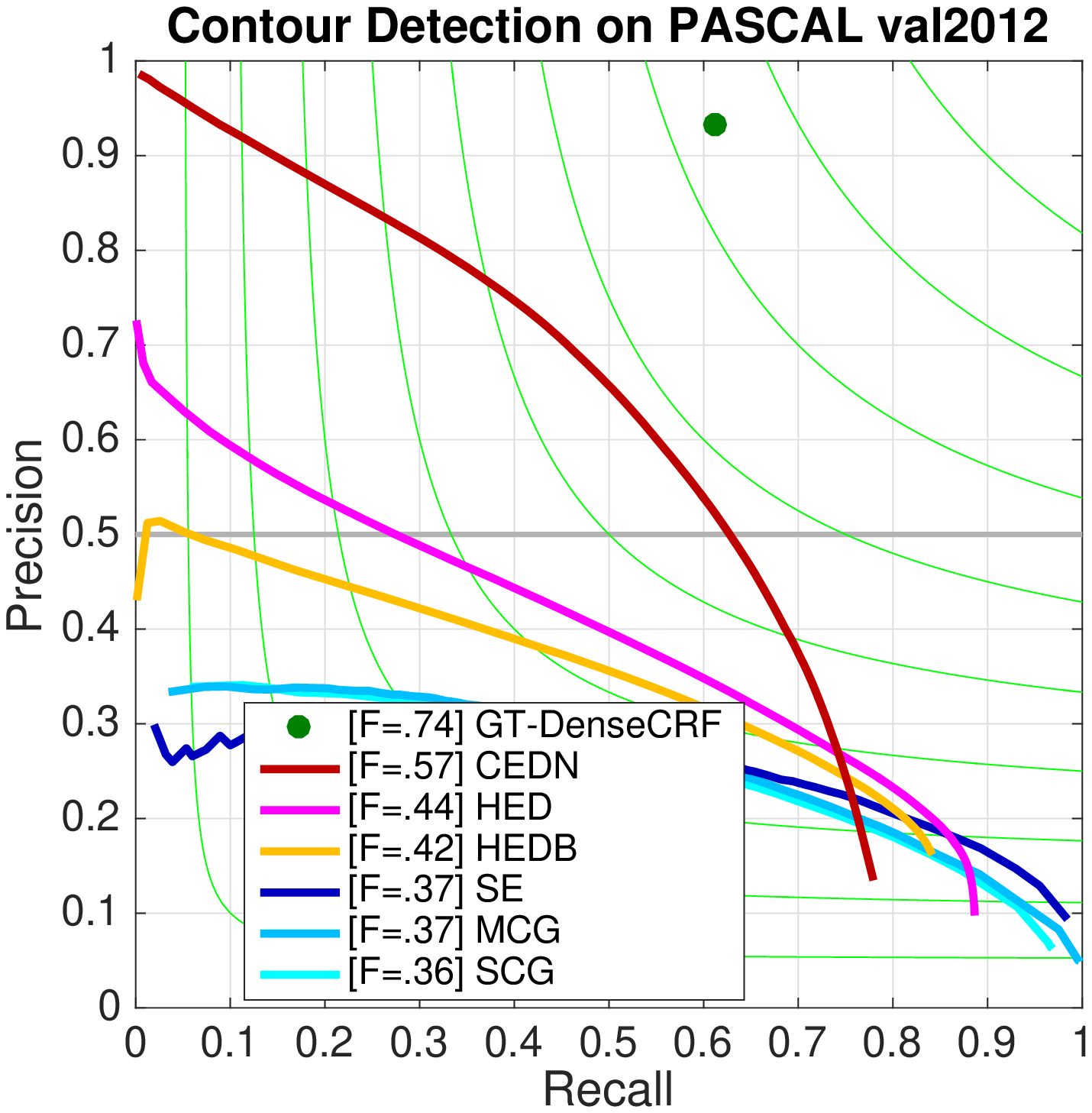}
\includegraphics[width=0.8\linewidth,keepaspectratio]{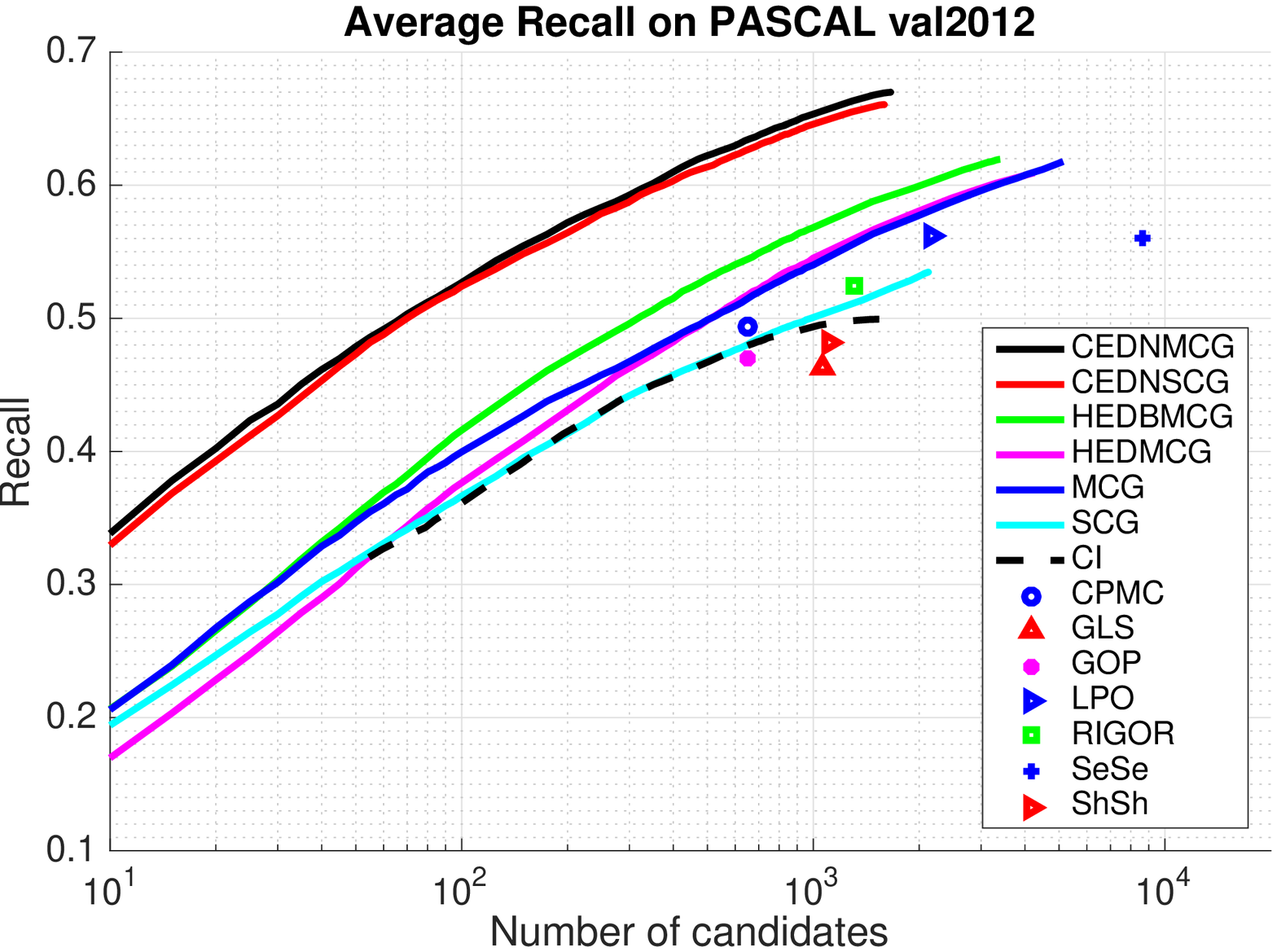}
\caption{Full comparisons with HED on PASCAL val2012.}
\label{fig:hed_compared}
\end{figure}

\paragraph{Comparing with HED trained on BSD and PASCAL VOC.}
We trained the HED model on PASCAL VOC using the same training data as our model with 30000 iterations. 
Its contour prediction precision-recall curve is illustrated in Figure~\ref{fig:hed_compared} with comparisons to our CEDN model, the pre-trained HED model on BSDS (referred as HEDB) and others. 
It can be seen that the F-score of HED is improved (from {\color{red} 0.42} to {\color{red} 0.44}) by training on PASCAL VOC  but still significantly lower than CEDN ({\color{red} 0.57}). 
We will present visual examples in the revised submission.
We further feed the HED edge maps into MCG for generating proposals and compare their average recalls with others in Figure~\ref{fig:hed_compared}.
We refer the results from PASCAL-trained HED as HEDMCG and the ones from pre-trained HED on BSDS as HEDBMCG. 
Unfortunately, the HEDMCG does not improve upon HEDBMCG.
With 1000 proposals, the average recalls of CEDNMCG, HEDMCG and HEDBMCG are about {\color{red}0.65}, {\color{red}0.57} and {\color{red}0.55}, respectively.

\end{document}